\documentclass[11pt]{article}

\usepackage[preprint]{acl}
\usepackage{times}
\usepackage{latexsym}
\usepackage{inconsolata}

\usepackage{threeparttable}
\usepackage{adjustbox}
\usepackage{amsmath}
\usepackage{amssymb}
\usepackage{multirow}
\usepackage{tcolorbox}
\usepackage{verbatim}
\usepackage{multicol}

\usepackage{pifont}
\usepackage{xcolor}
\definecolor{softgreen}{HTML}{77DD77}
\definecolor{softorange}{HTML}{FFB347}

\usepackage[utf8]{inputenc} % allow utf-8 input
\usepackage[T1]{fontenc}    % use 8-bit T1 fonts
\usepackage{hyperref}       % hyperlinks

\usepackage{url}            % simple URL typesetting
\usepackage{booktabs}       % professional-quality tables
\usepackage{amsfonts}       % blackboard math symbols
\usepackage{nicefrac}       % compact symbols for 1/2, etc.
\usepackage{microtype}      % microtypography
\usepackage{graphicx}
\usepackage{colortbl}
\newcommand{\highlight}[1]{%
  \cellcolor{cyan!#1!white}{#1}%
}

\usepackage{xspace}
\usepackage{gradient-text}
\newcommand{\multifinben}{\gradientRGB{MultiFinBen}{100,20,185}{240,0,15}\xspace}

\title{\textsc{\multifinben}: Benchmarking Large Language Models for Multilingual and Multimodal Financial Application}

\author{
\footnotesize
\textbf{Xueqing Peng\textsuperscript{1}},
\textbf{Lingfei Qian\textsuperscript{1}},
\textbf{Yan Wang\textsuperscript{1}},
\textbf{Ruoyu Xiang\textsuperscript{6}},
\textbf{Yueru He\textsuperscript{3}},
\textbf{Yang Ren\textsuperscript{1}},
\textbf{Mingyang Jiang\textsuperscript{1}},
\\
\footnotesize
\textbf{Vincent Jim Zhang\textsuperscript{1}},
\textbf{Yuqing Guo\textsuperscript{1}},
\textbf{Jeff Zhao\textsuperscript{1}},
\textbf{Huan He\textsuperscript{1}},
\textbf{Yi Han\textsuperscript{4}},
\textbf{Yun Feng\textsuperscript{17}},
\textbf{Yuechen Jiang\textsuperscript{7}},
\\
\footnotesize
\textbf{Yupeng Cao\textsuperscript{7}},
\textbf{Haohang Li\textsuperscript{7}},
\textbf{Yangyang Yu\textsuperscript{7}},
\textbf{Xiaoyu Wang\textsuperscript{6}},
\textbf{Penglei Gao\textsuperscript{8}},
\textbf{Shengyuan Lin\textsuperscript{20}},
\textbf{Keyi Wang\textsuperscript{3}},
\\
\footnotesize
\textbf{Shanshan Yang\textsuperscript{7}},
\textbf{Yilun Zhao\textsuperscript{2}},
\textbf{Zhiwei Liu\textsuperscript{10}},
\textbf{Peng Lu\textsuperscript{11}},
\textbf{Jerry Huang\textsuperscript{11}},
\textbf{Suyuchen Wang\textsuperscript{11}},
\textbf{Triantafillos Papadopoulos\textsuperscript{15,18}},
\\
\footnotesize
\textbf{Polydoros Giannouris\textsuperscript{10}},
\textbf{Efstathia Soufleri\textsuperscript{18}},
\textbf{Nuo Chen\textsuperscript{19}},
\textbf{Zhiyang Deng\textsuperscript{7}},
\textbf{Heming Fu\textsuperscript{1}},
\textbf{Yijia Zhao\textsuperscript{1}},
\textbf{Mingquan Lin\textsuperscript{13}},
\\
\footnotesize
\textbf{Meikang Qiu\textsuperscript{14}},
\textbf{Kaleb E. Smith\textsuperscript{5}},
\textbf{Arman Cohan\textsuperscript{2}},
\textbf{Xiao-Yang Liu\textsuperscript{3}},
\textbf{Jimin Huang\textsuperscript{1}},
\textbf{Guojun Xiong\textsuperscript{12}},
\textbf{Alejandro Lopez-Lira\textsuperscript{9}},
\\
\footnotesize
\textbf{Xi Chen\textsuperscript{6}},
\textbf{Junichi Tsujii\textsuperscript{16}},
\textbf{Jian-Yun Nie\textsuperscript{11}},
\textbf{Sophia Ananiadou\textsuperscript{10,18}},
\textbf{Qianqian Xie\textsuperscript{1,*}}
\\[1em]
\footnotesize
\textsuperscript{1}\textit{The FinAI},
\textsuperscript{2}\textit{Yale University},
\textsuperscript{3}\textit{Columbia University},
\textsuperscript{4}\textit{Georgia Institute of Technology},
\textsuperscript{5}\textit{NVIDIA},
\\
\footnotesize
\textsuperscript{6}\textit{New York University},
\textsuperscript{7}\textit{Stevens Institute of Technology},
\textsuperscript{8}\textit{Quantitative Health Sciences, Cleveland Clinic},
\\
\footnotesize
\textsuperscript{9}\textit{University of Florida},
\textsuperscript{10}\textit{University of Manchester},
\textsuperscript{11}\textit{University of Montreal},
\textsuperscript{12}\textit{Harvard University},
\\
\footnotesize
\textsuperscript{13}\textit{University of Minnesota},
\textsuperscript{14}\textit{Augusta University},
\textsuperscript{15}\textit{Athens University of Economics and Business},
\\
\footnotesize
\textsuperscript{16}\textit{National Institute of Advanced Industrial Science and Technology},
\textsuperscript{17}\textit{Asian Development Bank},
\\
\footnotesize
\textsuperscript{18}\textit{Archimedes, Athena Research Center},
\textsuperscript{19}\textit{National University of Singapore},
\textsuperscript{20}\textit{Carnegie Mellon University}
\\[0.6em]
\small{
\footnotesize
  \textbf{Correspondence:} \href{mailto:xqq.sincere@gmail.com}{xqq.sincere@gmail.com}
}
}

\begin{document}
\maketitle
\begin{abstract}
% Real-world financial applications process information in multiple languages from reports, news, scanned documents, and meeting recordings, yet most existing evaluations of large language models in finance rely on single-language, text-only datasets with tasks already solved by current models. To address these gaps, we introduce \textsc{\multifinben}, the first expert-annotated multilingual (five languages) and multimodal (text, vision, audio) benchmark designed to assess language models in financial contexts. It introduces the first financial multilingual tasks, which test expert-level analysis by connecting evidence across multiple languages on filings and news, together with Optical Character Recognition (OCR) tasks over scanned financial documents with tables and charts. Instead of aggregating all datasets, we filter a broad collection by the performance of advanced models, ensuring a balanced difficulty range and removing repeating tasks that offer little challenge. Evaluations of 21 LLMs show that even multimodal models like GPT-4o reaches only 46.01\%, performing limited on text but higher on vision and audio, with performance dropping by over ten points in multilingual settings. These results reveal persistent limitations in multilingual, multimodal, and expert-level financial tasks. All resources are publicly available. %\footnote{Code: \url{xxx}}.

Real-world financial analysis involves information across multiple languages and modalities, from reports and news to scanned filings and meeting recordings. Yet most existing evaluations of LLMs in finance remain text-only, monolingual, and largely saturated by current models.
To bridge these gaps, we present \textsc{\multifinben}, the first expert-annotated multilingual (five languages) and multimodal (text, vision, audio) benchmark for evaluating LLMs in realistic financial contexts.
\textsc{\multifinben} introduces two new task families: multilingual financial reasoning, which tests cross-lingual evidence integration from filings and news, and financial OCR, which extracts structured text from scanned documents containing tables and charts.
Rather than aggregating all available datasets, we apply a structured, difficulty-aware selection based on advanced model performance, ensuring balanced challenge and removing redundant tasks.
Evaluating 21 leading LLMs shows that even frontier multimodal models like GPT-4o achieve only 46.01\% overall, stronger on vision and audio but dropping sharply in multilingual settings.
These findings expose persistent limitations in multilingual, multimodal, and expert-level financial reasoning.
All datasets, evaluation scripts, and leaderboards are publicly released.
\footnote{Data:\hspace*{0.4em}\url{https://huggingface.co/datasets/TheFinAI/PolyFiQA-Easy}
\\\hspace*{4.4em}\url{https://huggingface.co/datasets/TheFinAI/PolyFiQA-Expert}
\\\hspace*{2.5em}\url{https://huggingface.co/datasets/TheFinAI/MultiFinBen-SpanishOCR}
\\\hspace*{2.5em}\url{https://huggingface.co/datasets/TheFinAI/MultiFinBen-EnglishOCR}.
\\Code: \url{https://github.com/xueqingpeng/MultiFinBen}}

% \footnote{Code: \url{https://anonymous.4open.science/r/MultiFinBen-DCF3/}}

% Recent advances in large language models (LLMs) have accelerated progress in financial NLP and applications, yet existing benchmarks remain limited to monolingual and unimodal settings, often over-relying on simple tasks and failing to reflect the complexity of real-world financial tasks. We introduce \textsc{\multifinben}, the first multilingual and multimodal benchmark tailored to the global financial domain, evaluating LLMs across modalities (text, vision, audio) and linguistic settings (monolingual, bilingual, multilingual) on domain-specific tasks. 
% We introduce two novel tasks, including PolyFiQA-Easy and PolyFiQA-Expert, the first multilingual financial benchmarks requiring models to perform complex reasoning over mixed-language inputs; and EnglishOCR and SpanishOCR, the first OCR-embedded financial QA tasks challenging models to extract and reason over information from visual-text financial documents.
% Moreover, we propose a dynamic, difficulty-aware selection mechanism and curate a compact, balanced benchmark rather than simple aggregation existing datasets. Extensive evaluations of 22 state-of-the-art models reveal that even the strongest models, despite their general multimodal and multilingual capabilities, struggle dramatically when faced with complex cross-lingual and multimodal tasks in financial domain. \textsc{\multifinben} is publicly released to foster transparent, reproducible, and inclusive progress in financial studies and applications. \footnote{Code: \url{xxx}}

\end{abstract}

\begin{figure}[h]
  \includegraphics[width=0.9\columnwidth]{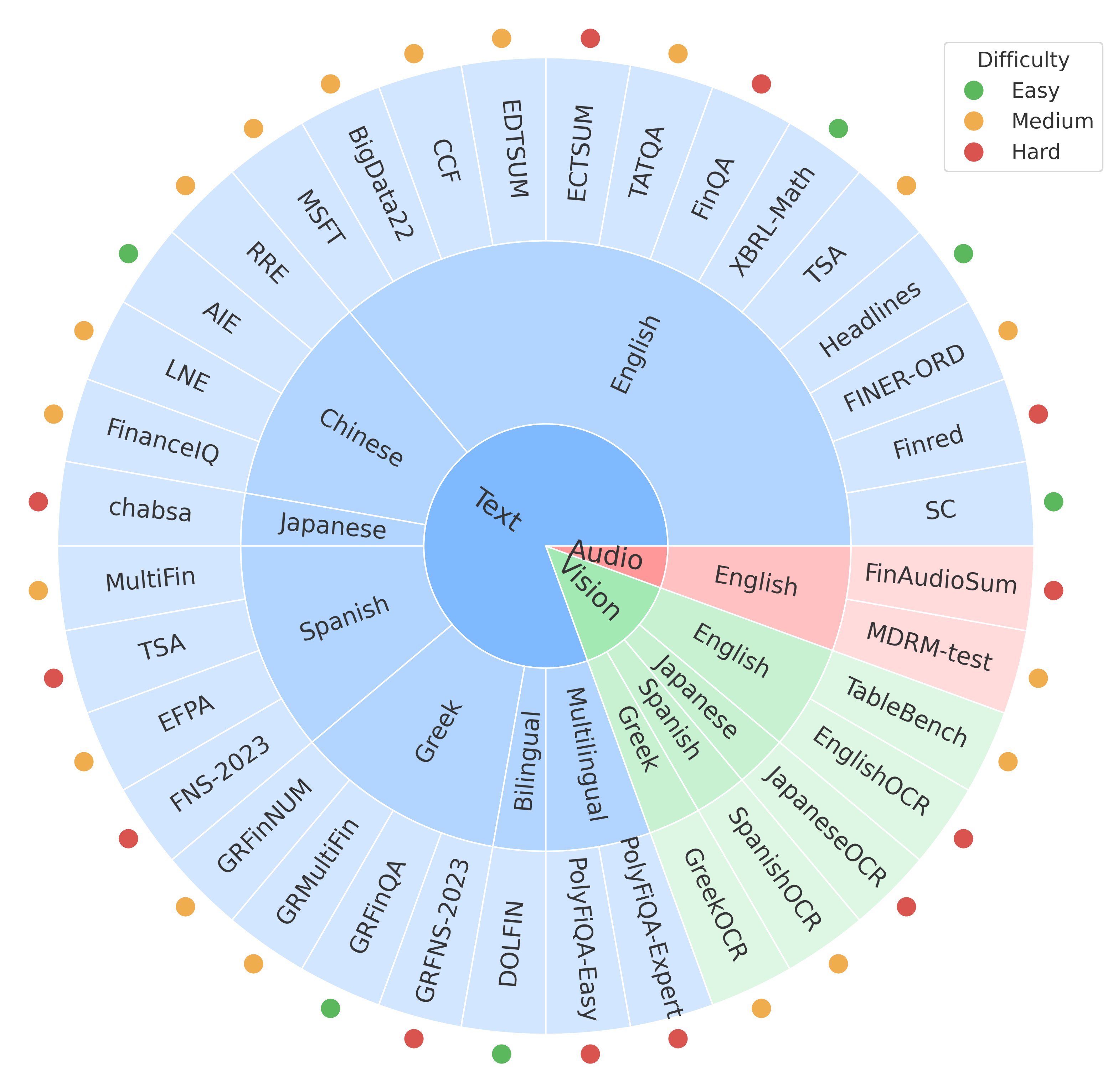}
  \caption{Overview of \textsc{\multifinben}.}
  \label{fig:fig_datasets}
\end{figure}

\begin{table}[h]
  \centering
  \scriptsize
  \setlength{\tabcolsep}{2pt}
  \begin{adjustbox}{max width=\columnwidth}
  \begin{tabular}{@{}lccccc|ccccc|c@{}}
    \toprule
    \textbf{Benchmark} & \textbf{EN} & \textbf{ZH} & \textbf{JA} & \textbf{ES} & \textbf{EL} & \textbf{BI} & \textbf{MU} & \textbf{Vis} & \textbf{OCR} & \textbf{Au} & \textbf{DA} \\
    \midrule
    % DOCMATH~\cite{zhao2023docmath}                                  & \ding{51} & \ding{55} & \ding{55} & \ding{55} & \ding{55} & \ding{55} & \ding{55} & \ding{55} & \ding{55} & \ding{55} & \ding{55} \\
    Plutus \cite{peng2025plutus}                                    & \ding{55} & \ding{55} & \ding{55} & \ding{55} & \ding{51} & \ding{55} & \ding{55} & \ding{55} & \ding{55} & \ding{55} & \ding{55} \\  
    CFinBench~\cite{nie2024cfinbenchcomprehensivechinesefinancial}  & \ding{55} & \ding{51} & \ding{55} & \ding{55} & \ding{55} & \ding{55} & \ding{55} & \ding{55} & \ding{55} & \ding{55} & \ding{55} \\
    FLARE-ES~\cite{zhang2024dolares}                                & \ding{51} & \ding{55} & \ding{55} & \ding{51} & \ding{55} & \ding{55} & \ding{55} & \ding{55} & \ding{55} & \ding{55} & \ding{55} \\
    FinBen~\cite{xie2024finbenholisticfinancialbenchmark}           & \ding{51} & \ding{55} & \ding{55} & \ding{51} & \ding{55} & \ding{55} & \ding{55} & \ding{55} & \ding{55} & \ding{55} & \ding{55} \\
    PIXIU~\cite{xie2023pixiu}                                       & \ding{51} & \ding{55} & \ding{55} & \ding{55} & \ding{55} & \ding{55} & \ding{55} & \ding{55} & \ding{55} & \ding{55} & \ding{55} \\
    DOLFIN~\cite{nakhlé2025dolfindocumentlevelfinancial}            & \ding{51} & \ding{55} & \ding{55} & \ding{51} & \ding{55} & \ding{51} & \ding{55} & \ding{55} & \ding{55} & \ding{55} & \ding{55} \\
    FinMME~\cite{luo-etal-2025-finmme}                              & \ding{51} & \ding{55} & \ding{55} & \ding{55} & \ding{55} & \ding{55} & \ding{55} & \ding{51} & \ding{55} & \ding{55} & \ding{55} \\
    MME-Finance~\cite{gan2024mmefinancemultimodalfinancebenchmark}  & \ding{51} & \ding{51} & \ding{55} & \ding{55} & \ding{55} & \ding{55} & \ding{55} & \ding{51} & \ding{55} & \ding{55} & \ding{55} \\
    FinAudio~\cite{cao2025finaudiobenchmarkaudiolarge}              & \ding{51} & \ding{55} & \ding{55} & \ding{55} & \ding{55} & \ding{55} & \ding{55} & \ding{55} & \ding{55} & \ding{51} & \ding{55} \\
    \textsc{\multifinben} \textbf{(ours)}                            & \ding{51} & \ding{51} & \ding{51} & \ding{51} & \ding{51} & \ding{51} & \ding{51} & \ding{51} & \ding{51} & \ding{51} & \ding{51} \\
    \bottomrule
  \end{tabular}
  \end{adjustbox}
  \caption{\small Comparison with existing benchmarks. EN: English, ZH: Chinese, JA: Japanese, ES: Spanish, EL: Greek, BI: Bilingual, MU: Multilingual, Vis: Chart/Tabular, OCR: Optical characeter recognition, Au: Audio, DA: Difficulty-aware.}
  \label{tab:tab_relatedwork_abbr}
\end{table}

\begin{figure}[h]
  \includegraphics[width=\columnwidth]{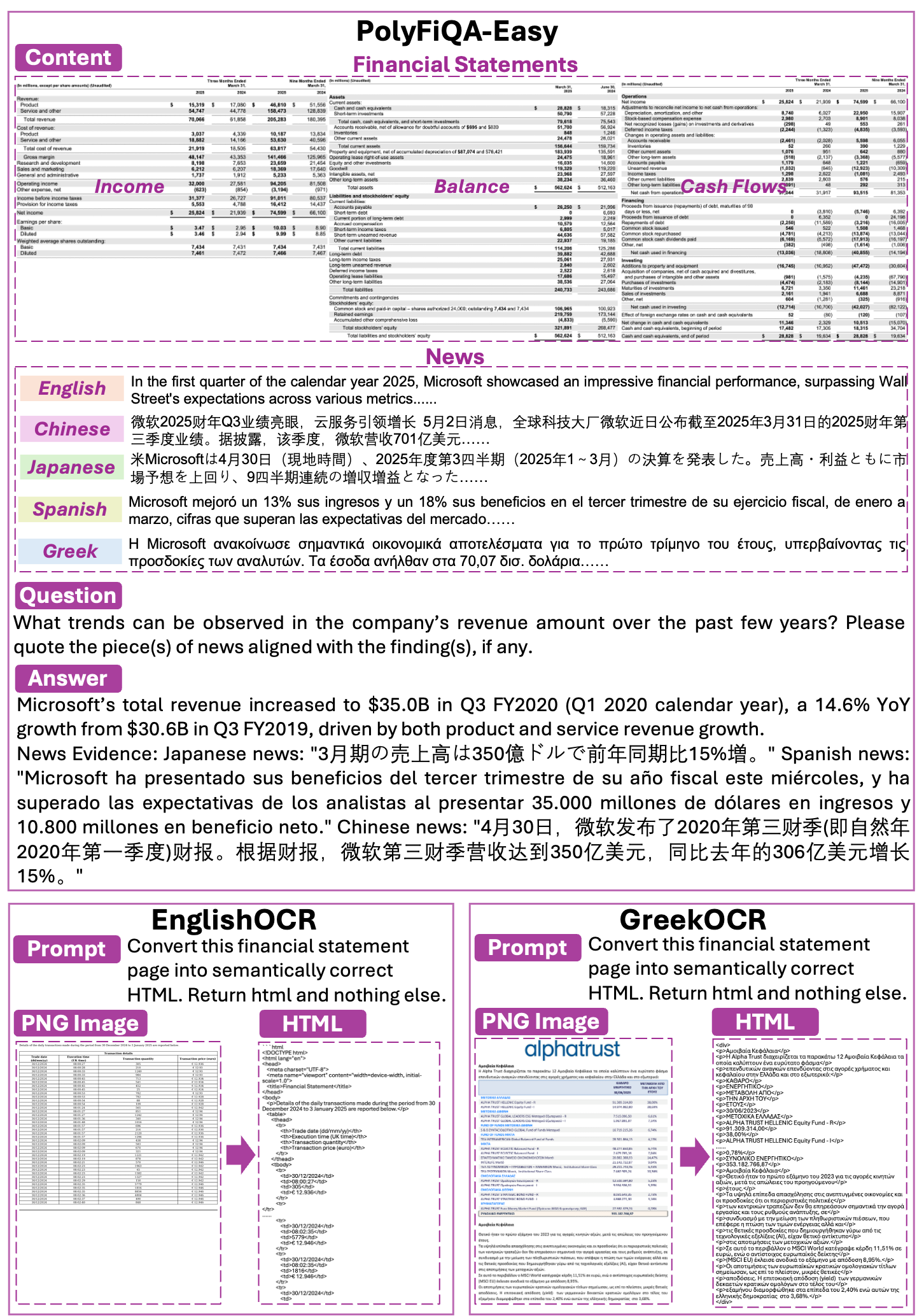}
  \caption{Representation examples of \textit{PolyFiQA-Easy}, \textit{EnglishOCR}, and \textit{GreekOCR}.}
  \label{fig:fig_noveldata}
\end{figure}
\vspace{-10pt}

\section{Introduction}
\label{sec:Introduction}
LLMs have demonstrated remarkable prowess in various tasks~\cite{zhao2025mmvumeasuringexpertlevelmultidiscipline, Chen_2025}, yet their applications to finance~\cite{Liu2023FinGPT,xie2023pixiu,wu2023bloomberggpt} require the continual push of the limits of model capabilities~\cite{xiong2025flagtraderfusionllmagentgradientbased}. Although there are already several specialized benchmarks~\cite{xie2024finben, li2024investorbenchbenchmarkfinancialdecisionmaking, cao2025finaudiobenchmarkaudiolarge} that evaluate LLM in core financial tasks, they still suffer from three critical limitations: 
% First, they are overwhelmingly \textbf{monolingual and monomodal}. However, real-world financial applications involve mixed inputs that require multilingual semantic parsing, multimodal inputs of text, tables, charts, and audio, and cross-cultural contextual understanding.
% %Secondly, existing benchmarks are largely \textbf{static}. Datasets like FinanceBench have not been updated since 2021, allowing state-of-the-art models to routinely exceed 98\% accuracy on repetitive question templates. This saturation reduces benchmarks' discriminative power and increases evaluation costs without providing meaningful insights.
% Secondly, existing benchmarks rely on \textbf{simple aggregation} without difficulty-aware selection, leading to over-weighting of easy and duplicated tasks. For example, in FinBen~\cite{xie2024finben}, 8 of 36 datasets are textual analysis tasks, with 7 being simple enough for zero-shot LLMs to exceed 60\% accuracy, causing inflated overall scores and increased evaluation costs while failing to expose model weaknesses and providing meaningful insights.

\begin{itemize}
    \item \textit{Monolingual and monomodal scope.} Prior benchmarks largely focus on English text only, whereas real-world financial applications demand cross-lingual understanding and multimodal reasoning across text, tables, charts, PDFs, and audio.
    \item \textit{Imbalanced task difficulty.} Existing benchmarks are typically constructed by simple aggregation without difficulty calibration. This leads to an overrepresentation of trivial tasks. For example, in FinBen \cite{xie2024finben}, 7 out of 8 text classification datasets can already be solved above 60\% accuracy in zero-shot, resulting in inflated scores and obscuring true model weaknesses.
    \item \textit{Missing key financial scenarios.} No prior benchmark addresses multilingual financial reasoning that integrates heterogeneous sources, nor OCR tasks that require extracting and reasoning over scanned financial PDFs. These scenarios are ubiquitous in practice yet entirely absent in prior benchmarks.
\end{itemize}

\paragraph{Benchmark Design.} 
To address these gaps, we introduce \textsc{\multifinben}, the first unified benchmark that spans three modalities (text, vision, and audio), three linguistic settings (monolingual, bilingual, and multilingual), and seven task categories across three difficulty tiers. In total, it comprises 36 datasets in five languages (English, Chinese, Japanese, Spanish, and Greek). 
Unlike prior benchmarks limited to monolingual or unimodal corpora, \textsc{\multifinben} introduces two novel multilingual financial reasoning datasets, \textit{PolyFiQA-Easy} and \textit{PolyFiQA-Expert}, requiring joint reasoning over mixed-language inputs. In addition, we present the first financial OCR datasets (\textit{EnglishOCR}, \textit{JapaneseOCR}, \textit{SpanishOCR}, and \textit{GreekOCR}) to evaluate models on document-image understanding, a setting ubiquitous in practice but overlooked in prior work. Finally, we propose a difficulty-aware selection mechanism that retains one dataset per modality–language–task tier with the largest inter-model gap, avoiding redundancy and ensuring meaningful evaluation. 

\paragraph{Evaluation and Findings.} 
We benchmark 21 frontier models on \textsc{\multifinben}. Even the strongest model, GPT-4o, achieves only 46.01\%, while monolingual and unimodal models fail dramatically on unsupported inputs, underscoring the necessity of cross-modal reasoning. Its significant low performance in multilingual scenario (9.79\% and 5.31\%) highlights persistent weaknesses in cross-lingual generalization. Our difficulty-aware design reveals steep drops from 32.73\% (easy) to 7.20\% (hard), exposing the gap between current model capabilities and real-world complexity. Crucially, our newly introduced datasets (\textit{PolyFiQA-Easy/Expert}, \textit{EnglishOCR}, and \textit{JapaneseOCR}) emerge as among the hardest challenges, demonstrating how \textsc{\multifinben} systematically surfaces limitations and provides a roadmap for developing more realistic financial evaluation settings.

\paragraph{Our Contributions.} 
We summarize the main contributions of this work as follows:

 $\triangleright$ \textit{First multilingual and multimodal benchmark for financial LLMs.} \textsc{\multifinben} spans five languages (EN, ZH, JA, ES, EL), three modalities (text, vision, audio), and seven task categories, offering the most comprehensive evaluation setting.
    
    $\triangleright$ \textit{Novel and realistic datasets.} We introduce \textit{PolyFiQA-Easy} and \textit{PolyFiQA-Expert}, the first multilingual financial QA datasets requiring cross-lingual reasoning, as well as \textit{EnglishOCR}, \textit{JapaneseOCR}, \textit{SpanishOCR}, and \textit{GreekOCR}, the first financial OCR tasks targeting scanned financial documents.
    
    $\triangleright$ \textit{Difficulty-aware evaluation framework.} We propose a principled dataset selection mechanism based on inter-model performance gaps, ensuring balanced coverage across easy to hard tiers, in contrast to prior aggregation-based benchmarks.
    
    $\triangleright$  \textit{Comprehensive model evaluation and insights.} We evaluate 21 state-of-the-art models and reveal substantial performance gaps across modalities, languages, and difficulty levels, with even GPT-4o achieving only 46.01\%. This demonstrates the necessity of our benchmark for stress-testing current models and guiding future development.
    
   $\triangleright$  \textit{Open resources.} We release all datasets and code to foster transparency, reproducibility, and future extensions by the research community.

% % results
% We evaluate 22 frontier models on \textsc{\multifinben}. Overall, GPT-4o leads with 50.67\%, while monomodal and monolingual models lag far behind. Unimodal models fail on unsupported inputs, underscoring the necessity of cross-modal reasoning. A 10.29 pp gap between multilingual (7.50\%) and monolingual (17.79\%) QA reveals that existing models still struggle with cross-lingual generalization in financial contexts. Our structured difficulty design exposes steep performance drops from 31.24\% (easy) to 6.63\% (hard), reflecting the gap between current model capabilities and real-world financial task complexity. Crucially, newly introduced datasets, such as PolyFiQA-Easy/Expert and the first financial OCR QA task, surface as among the hardest challenges, addressing modality and linguistic gaps overlooked by prior benchmarks. These results demonstrate that \textsc{\multifinben} not only reveals current model limitations in cross-modal, cross-lingual, and domain-specific applications but also provides a systematic framework to guide future model improvements and the development of harder and more realistic financial datasets.

\section{\textsc{\multifinben} Benchmark}

In this section, we present \textsc{\multifinben}, the first unified benchmark for evaluating LLMs in the financial domain across diverse modalities, linguistic settings, and tasks (Table ~\ref{tab:tab_datasets}).
%Beyond simple aggregation, \textsc{\multifinben} introduces a structured, difficulty-aware design that enables resource-efficient, scalable, and fine-grained evaluation of multilingual and multimodal financial reasoning.
%Unlike prior financial benchmarks that aggregate heterogeneous datasets without calibration, \textsc{\multifinben} is designed under three principles: (i) balanced difficulty, (ii) cross-lingual coverage, and (iii) modality realism.
%This integration yields a balanced, multilingual, and multimodal benchmark that enables efficient, comprehensive and fine-grained evaluation across the global financial domain.

\begin{table*}[h]
\centering
\small
\setlength{\tabcolsep}{4pt}
\setlength{\itemsep}{-10pt}
\setlength{\parsep}{0pt}
\setlength{\parskip}{0pt}
\renewcommand{\arraystretch}{0.95}

\begin{threeparttable}
\begin{adjustbox}{width=\textwidth}
\begin{tabular}{@{}llcccrcccc@{}}
\toprule
\textbf{Modality} & \textbf{Language} & \textbf{Task} & \textbf{Dataset} & \textbf{Source} & \textbf{Size} & \textbf{Metric} & \textbf{License} & \textbf{Difficulty} \\
\midrule
\addlinespace[0.8ex]
\multirow{29}{*}{Text} & \multirow{13}{*}{English} & IE & SC \cite{mariko-etal-2020-financial} & Financial News & 8{,}630 & F1 & CC BY 4.0 & \textcolor{softgreen}{Easy} \\
& & IE & FinRED \cite{sharma2022finred} & Webhose Financial News and Earning Call Transcrip & 1{,}068 & F1 & Public & \textcolor{red}{Hard} \\
& & IE & FINER-ORD \cite{shah2023finer} & Financial News & 1{,}080 & Entity F1 & CC BY-NC 4.0 & \textcolor{softorange}{Medium} \\
& & TA & Headlines \cite{sinha2021impact} & Financial News & 2{,}283 & Avg F1 & CC BY-SA 3.0 & \textcolor{softgreen}{Easy} \\
& & TA & TSA \cite{cortis2017semeval} & Financial Microblog & 561 & Accuracy & CC BY-NC-SA 4.0 & \textcolor{softorange}{Medium} \\
& & QA & XBRL-Math \cite{chen2022convfinqa} & XBRL Agent & 1{,}000 & Accuracy & CDLA-Permissive 2.0 & \textcolor{softgreen}{Easy} \\
& & QA & FinQA \cite{chen2021finqa} & Annual Reports & 1,147 & Accuracy & MIT License & \textcolor{red}{Hard} \\
& & QA & TATQA \cite{zhu2021tat} & Annual Reports & 1{,}668 & Accuracy & MIT License & \textcolor{softorange}{Medium} \\
& & TG & ECTSUM \cite{mukherjee2022ectsum} & Earnings Call Transcripts from the Motley Fool & 495 & ROUGE-1 & Public & \textcolor{red}{Hard} \\
& & TG & EDTSUM \cite{xie2023pixiu} & Financial News & 2{,}000 & ROUGE-1 & Public & \textcolor{softorange}{Medium} \\
% & & RM & polish & Emerging Markets Information Service\tnote{1} & 1{,}736 & MCC & CC BY 4.0 & \textcolor{softorange}{Medium} \\
& & RM & CCF \cite{feng2024empowering} & Credit Card Transactions & 2{,}278 & MCC & DbCL v1.0 & \textcolor{softorange}{Medium} \\
& & FO & BigData22 \cite{soun2022accurate} & Tweets, Historical Prices & 1,470 & MCC & Public & \textcolor{softorange}{Medium} \\
% & & DM & NFLX & TradeTheEvent Dataset & 270 & SR & Open Source & \textcolor{softorange}{Medium} \\
& & DM & MSFT \cite{yu2024finconsynthesizedllmmultiagent} & TradeTheEvent Dataset & 222 & SR & Open Source & \textcolor{softorange}{Medium} \\
\cmidrule(lr){2-9}
& \multirow{4}{*}{Chinese} & IE & RRE \cite{jiajia-etal-2024-auditwen} & Regulatory Documents & 117 & Accuracy & CC BY-NC & \textcolor{softorange}{Medium} \\
& & TA & AIE \cite{jiajia-etal-2024-auditwen} & Regulatory Documents & 1{,}573 & Accuracy & CC BY-NC & \textcolor{softgreen}{Easy} \\
% & & TA & APE & Regulatory Documents & 153 & Accuracy & CC BY-NC & \textcolor{softorange}{Medium} \\
& & TA & LNE \cite{jiajia-etal-2024-auditwen} & Regulatory Documents & 218 & Accuracy & CC BY-NC & \textcolor{softorange}{Medium} \\
& & QA & FinanceIQ \cite{10.1145/3583780.3615285} & Exam Questions & 7{,}123 & Accuracy & CC BY-NC-SA 4.0 & \textcolor{softorange}{Medium} \\
\cmidrule(lr){2-9}
& \multirow{1}{*}{Japanese} & TA & chabsa \cite{kubo2018chabsa} & Securities Reports & 7{,}723 & Macro F1 & CC BY 4.0 & \textcolor{red}{Hard} \\
\cmidrule(lr){2-9}
& \multirow{4}{*}{Spanish} & TA & MultiFin \cite{jorgensen2023multifin} & Article Headlines & 368 & Accuracy & MIT License & \textcolor{softorange}{Medium} \\
& & TA & TSA \cite{pan2023evaluation} & News Headlines & 726 & Accuracy & Public & \textcolor{red}{Hard} \\
% & & TA & FinanceES & News Headlines & 1{,}621 & Accuracy & Public & \textcolor{red}{Hard} \\
% & & QA & EFP & Exam Questions & - & Accuracy & Public & \textcolor{softorange}{Medium} \\
& & QA & EFPA \tnote{5} & Exam Questions & 50 & Accuracy & Public & \textcolor{softorange}{Medium} \\
& & TG & FNS-2023 \tnote{6} & Annual Reports & 50 & ROUGE-1 & Public & \textcolor{red}{Hard} \\
\cmidrule(lr){2-9}
& \multirow{4}{*}{Greek} & IE & GRFinNUM \cite{peng2025plutus} & Annual Reports & 100 & Entity F1 & Public & \textcolor{softorange}{Medium} \\
% & & IE & GRFinNER & Annual Reports\tnote{3} & 100 & Entity F1 & Public & \textcolor{softorange}{Medium} \\
& & TA & GRMultiFin \cite{jorgensen-etal-2023-multifin} & Article Headlines & 54 & Accuracy & CC BY-NC 4.0 & \textcolor{softorange}{Medium} \\
& & QA & GRFinQA \cite{peng2025plutus} & Exam Questions & 225 & Accuracy & Public & \textcolor{softgreen}{Easy} \\
& & TG & GRFNS-2023 \cite{10386228} & Annual Reports & 50 & ROUGE-1 & CC BY 4.0 & \textcolor{red}{Hard} \\
\cmidrule(lr){2-9}
& \multirow{1}{*}{Bilingual} & TG & DOLFIN \cite{nakhlé2025dolfindocumentlevelfinancial} & Fundinfo Financial Documents & 1{,}932 & Comet-da-22 & MIT License & \textcolor{softgreen}{Easy} \\
\cmidrule(lr){2-9}
& \multirow{2}{*}{Multilingual} & QA & \textbf{PolyFiQA-Easy} & Financial Reports\tnote{1} and News\tnote{7} & 172 & ROUGE-1 & Public & \textcolor{red}{Hard} \\
& & TG & \textbf{PolyFiQA-Expert} & Financial Reports\tnote{1} and News\tnote{7} & 172 & ROUGE-1 & Public & \textcolor{red}{Hard} \\
\midrule
\multirow{5}{*}{Vision} & \multirow{2}{*}{English} & IE & \textbf{EnglishOCR} & SEC EDGAR Company Filings\tnote{1} & 7{,}961 & ROUGE-1 & Public & \textcolor{red}{Hard} \\
% & & QA & ChartBench & Unichart, Chart2Text, and ChartQA & 350 & Accuracy & Public & \textcolor{softorange}{Medium} \\
& & QA & TableBench \cite{xie2024open} & SynthTabNet (Fintabnet and Marketing categories) & 450 & Accuracy & Public & \textcolor{softorange}{Medium} \\
\cmidrule(lr){2-9}
& \multirow{1}{*}{Japanese} & IE & \textbf{JapaneseOCR} & FSA White Paper\tnote{2} & 17{,}586 & ROUGE-1 & Public & \textcolor{red}{Hard} \\
\cmidrule(lr){2-9}
& \multirow{1}{*}{Spanish} & IE & \textbf{SpanishOCR} & Regulatory Documents\tnote{3} & 12{,}819 & ROUGE-1 & Public & \textcolor{softorange}{Medium} \\
\cmidrule(lr){2-9}
& \multirow{1}{*}{Greek} & IE & \textbf{GreekOCR} & Annual Company Filings on Athens Stock Exchange\tnote{4} & 6{,}533 & ROUGE-1 & Public & \textcolor{softorange}{Medium} \\
\midrule
\multirow{2}{*}{Audio} & \multirow{2}{*}{English} & TG & MDRM-test \cite{qin2019you} & Earnings Conference Calls & 22{,}208 & WER & Public & \textcolor{softorange}{Medium} \\
% & & TG & SPGISpeech-test  & Earnings Conference Calls & 39{,}341 & WER & Public & \textcolor{softorange}{Medium} \\
% & & TG & Earning-21 & Earnings Conference Calls & 44 & WER & Public & \textcolor{softorange}{Medium} \\
% & & TG & Earning-22 & Earnings Conference Calls & 125 & WER & Public & \textcolor{softorange}{Medium} \\
& & TG & FinAudioSum \cite{cao2025finaudiobenchmarkaudiolarge} & Earnings Conference Calls & 64 & ROUGE-L & Public & \textcolor{red}{Hard} \\
\bottomrule
\end{tabular}
\end{adjustbox}

% \vspace{0.5em}
\begin{tablenotes}
\tiny
\item[1] \url{https://www.sec.gov/search-filings}
\item[2] \url{https://www.fsa.go.jp/en/}
\item[3] \url{https://www.bvl.com.pe/en/home-general}\\
\item[4] \url{https://www.athexgroup.gr/en/market-data/issuers}\hspace{-1.3em}
\item[5] \url{https://efpa-eu.org/}\hspace{-1.3em}
\item[6] \url{https://wp.lancs.ac.uk/cfie/fns2023/}\hspace{-1.3em}
\item[7] Please check Appendix \ref{sec:sec_news_data_source} for more details.
\end{tablenotes}
\end{threeparttable}
\caption{Overview of \textsc{\multifinben}.}
\label{tab:tab_datasets}
\end{table*}

\subsection{Overview}

Built upon this foundation, \textsc{\multifinben} systematically organizes datasets along three structural dimensions (i.e., modality, language, and task) to enable comprehensive and interpretable evaluation. Rather than aggregating heterogeneous resources, it introduces a difficulty-aware benchmarking framework that selects representative datasets based on their discriminative power across models, ensuring balanced coverage and meaningful scalability.
To address gaps in existing resources, we further design new datasets and tasks that capture underrepresented yet crucial financial scenarios.
Together, these principles establish \textsc{\multifinben} as a unified, rigorous, and extensible framework for assessing financial reasoning across modalities, languages, and task types.

\paragraph{Modalities.}
Beyond text, it uniquely integrates visual (charts, tables, images) and audio (earnings calls) modalities into a unified financial benchmark, reflecting common formats of financial dissemination and enabling realistic assessment of multimodal understanding in financial contexts.

\paragraph{Linguistics.}
\textsc{\multifinben} is the first to evaluate LLMs under three linguistic settings (monolingual, bilingual, and multilingual) covering five typologically and economically diverse languages (English, Chinese, Japanese, Spanish, and Greek).
This design spans major global financial regions, providing a realistic setting for evaluating cross-lingual financial understanding.

% In \textsc{\multifinben}, we are the first to evaluate financial LLMs across three linguistic settings: \textbf{monolingual}, \textbf{bilingual}, and \textbf{multilingual}. %In the monolingual setting, models operate within a single language; the bilingual setting involves cross-lingual transferring; and the multilingual setting requires processing across multiple languages simultaneously.
% To support this, we include five typologically and economically diverse languages (English, Chinese, Japanese, Spanish, and Greek), balancing linguistic diversity, writing systems, and global financial regions across North America, East Asia, Southern Europe, and Latin America.

% In addition to text, we are also the first to include both \textbf{visual} and \textbf{audio} modalities in a unified financial benchmark for evaluating LLMs.
% In the visual modality, we focus on charts, tabular data, and text embedded in images, reflecting formats commonly found in financial reports and regulatory documents.
% In the audio modality, we include spoken financial content such as earnings calls, which contain spontaneous speech, domain-specific terminology, and prosodic cues that are essential for understanding intent, sentiment, and narrative context~\cite{qin2019you, cao2024ecc}.
% %By evaluating models on both modalities, we enable a more rigorous and realistic assessment of their ability to perform in high-stakes, information-rich financial environments.

\paragraph{Task Categories.}
Inspired by FinBen \cite{xie2024finbenholisticfinancialbenchmark}, our datasets are organized into seven core financial NLP tasks (Appendix~\ref{sec:sec_task_categories}): Information Extraction (IE), Textual Analysis (TA), Question Answering (QA), Text Generation (TG), Risk Management (RM), Forecasting (FO), and Decision-Making (DM).
This organization ensures broad task coverage and supports fine-grained evaluation of reasoning capabilities across diverse financial applications.

\subsection{Structured Difficulty-Aware Benchmarking}
Existing financial and cross-domain benchmarks often emphasize breadth, aggregating large numbers of datasets across tasks and modalities. While such aggregation maximizes coverage, it obscures the underlying sources of difficulty and makes model progress difficult to interpret.
In contrast, \textsc{\multifinben} adopts a structured, difficulty-aware design that disentangles evaluation along three orthogonal dimensions, i.e., modality, language, and task, enabling controlled comparison and clear attribution of performance to specific reasoning or linguistic factors~\cite{suzgun2022challengingbigbenchtaskschainofthought, glazer2024frontiermathbenchmarkevaluatingadvanced}.

To quantify dataset difficulty in a reproducible and model-agnostic manner, we compute the mean standardized performance of two representative large language models, GPT-4o~\cite{hurst2024gpt} and LLaMA-3.1-70B-Instruct~\cite{dubey2024llama}, defined as
\begin{equation}
    \bar{s}_d = \tfrac{1}{2}\big(s_{\text{GPT-4o},d} + s_{\text{LLaMA3.1},d}\big).
\end{equation}
These models are chosen for their complementary characteristics: GPT-4o provides frontier-level reasoning and instruction alignment, while LLaMA-3.1-70B-Instruct serves as an open, transparent baseline. Their differences in architecture, scale, and training corpus yield a balanced estimate of intrinsic dataset complexity, independent of individual model bias.

We categorize datasets into three tiers, \emph{easy} ($\bar{s}_d > 60$), \emph{medium} ($20 \le \bar{s}_d \le 60$), and \emph{hard} ($\bar{s}_d < 20$), corresponding to reliable competence, transitional reasoning, and consistent failure, respectively (Table~\ref{tab:finben-difficulty}; Figure~\ref{fig:finben-difficulty}; Appendix~\ref{sec:sec_benchmarking}).
Within each modality–language–task configuration, we retain the dataset exhibiting the largest inter-model divergence, as such cases most effectively reveal capability boundaries. In the event of ties, the lower-performing dataset is selected to preserve headroom for future progress.
This principled selection process yields a compact yet diagnostic benchmark that balances interpretability with challenge, supporting layered evaluation across easy, medium, and hard regimes while maintaining scalability and discriminative power.

\subsection{Benchmark Composition}

\textsc{\multifinben} is organized around two complementary sources of datasets, aligning with our three evaluation dimensions: modality, language, and task.
(\textit{i}) Newly introduced resources expand the benchmark’s task coverage, including two multilingual financial QA datasets (\textit{PolyFiQA-Easy} and \textit{PolyFiQA-Expert}; Section~\ref{sec:sec_polyfiqa}) and four OCR-based datasets for document-image understanding (\textit{EnglishOCR}, \textit{JapaneseOCR}, \textit{SpanishOCR}, and \textit{GreekOCR}; Section~\ref{sec:sec_ocr}).
(\textit{ii}) Existing financial benchmarks are integrated through our structured, difficulty-aware selection framework, ensuring balanced representation across text, vision, and audio modalities and across five languages.
Together, these resources define the complete task suite of \textsc{\multifinben}, supporting evaluation of financial reasoning, comprehension, and generation across multilingual and multimodal settings.

\paragraph{Text.}
In the monolingual setting, 26 datasets across five languages are selected from 68 available candidates following our difficulty calibration (Appendix~\ref{sec:sec_textual_monolingual_benchmarks}).
In the bilingual setting, we include the English–Spanish subset of \textit{DOLFIN}~\cite{nakhlé2025dolfindocumentlevelfinancial}, a document-level financial translation benchmark.
As no public datasets exist for multilingual financial reasoning, we introduce two new multilingual QA datasets, \textit{PolyFiQA-Easy} and \textit{PolyFiQA-Expert}, designed to evaluate cross-lingual comprehension and reasoning.

\paragraph{Vision.}
For the visual modality, we integrate two English QA datasets from Open-FinLLMs~\cite{xie2024open}.
Among these, \textit{TableBench} is retained due to its greater inter-model performance variance, offering richer signal for difficulty assessment.
We further contribute four newly constructed OCR-based datasets (Section~\ref{sec:sec_ocr}) to evaluate document-level text extraction and structural understanding across languages.

\paragraph{Audio.}
For the audio modality, we initially consider five English text-generation datasets from FinAudio~\cite{cao2025finaudiobenchmarkaudiolarge}.
After applying our structured filtering, two datasets are retained: \textit{MDRM}~\cite{qin2019you}, a medium-tier ASR task featuring short earnings call clips, and \textit{FinAudioSum}~\cite{cao2025finaudiobenchmarkaudiolarge}, a hard-tier summarization task comprising long-form financial recordings (Appendix~\ref{sec:sec_audio_benchmark}).

Together, these resources enable systematic evaluation of financial understanding across text, vision, and audio modalities under consistent difficulty calibration.

\subsection{Novel Tasks and Datasets}

To close critical gaps in multilingual and visual financial reasoning, we introduce two new task families within \textsc{\multifinben}:
(1) Multilingual Financial Question Answering (i.e., \textit{PolyFiQA}), the first task for multilingual cross-document reasoning grounded in native financial sources; and
(2) Financial OCR, the first OCR task for structured extraction from scanned financial documents.
Together, these tasks extend financial evaluation beyond text-based English corpora, bringing cross-lingual reasoning and document-level visual understanding into one unified framework.

\subsubsection{Multilingual Financial QA (\textit{PolyFiQA})}

\label{sec:sec_polyfiqa}
Unlike prior work relying on translation-based alignment, \textit{PolyFiQA} draws directly from native-language financial disclosures and contemporaneous news, preserving authentic phrasing, cultural framing, and domain complexity.
This design enables realistic evaluation of LLMs in real-world multilingual financial contexts, where decision-making increasingly spans languages and jurisdictions.

\paragraph{Task Definition.}

Provided a multilingual  context
 \[ C = \{ R, N_{\text{en}}, N_{\text{zh}}, N_{\text{ja}}, N_{\text{es}}, N_{\text{el}} \}, \] where \( R \) denotes a financial report (10-K or 10-Q) and \( N_{\text{lang}} \) represents contemporaneous news articles in five languages. 
Given a carefully designed natural language question \( q \) and the associated multilingual context \( C \), the model must generate an answer \( a \) grounded in integrated multilingual information.
This task challenges models to perform \textit{multilingual cross-document reasoning}, mirroring how analysts synthesize signals from diverse linguistic sources under real financial conditions.

\paragraph{Data Construction.}

We release two complementary datasets: \textit{PolyFiQA-Easy} and \textit{PolyFiQA-Expert}, targeting different reasoning depths.
Financial filings are collected from SEC EDGAR~\footnote{\url{https://www.sec.gov/search-filings}}
 and paired with temporally aligned multilingual news (Appendix~\ref{sec:sec_news_data_source}). To maintain focus and reduce noise, we extract three core statements from lengthy financial reports: Comprehensive Income, Consolidated Balance Sheets, and Cash Flows.
Low-resource cases are supplemented with expert-authored, native-language news verified by bilingual financial professionals (Appendix~\ref{sec:sec_news_generation}).

To ensure domain fidelity, we adopt an \emph{\textbf{expert-in-the-loop}} pipeline (Figure~\ref{fig:fig_polyfiqa}): three financial experts curated, annotated, and validated all data, spending over 130 hours in Label Studio (Appendix~\ref{sec:sec_annotation_process}).
Tier-specific guidelines (Appendix~\ref{sec:sec_annotator_guideline}) and iterative pilots ensured consistency.
Each question was authored and scored for two dimensions, i.e., Relevance (1–4) and Consistency (1–3), by multiple annotators following a structured validation protocol.
Only instances with cumulative scores above 5 were retained, yielding inter-annotator agreement above 89\% for both tiers.
This rigorous process results in a high-quality, auditable dataset for fine-grained evaluation of multilingual financial reasoning. More details can be found in Appendix \ref{sec:data_construction}.

\paragraph{Evaluation Metric.}

We adopt ROUGE-1~\cite{lin-2004-rouge} to measure unigram overlap between model predictions and references, offering a proxy for content coverage and factual alignment in multilingual QA and TG tasks.

\subsubsection{Optical Character Recognition (OCR)}
\label{sec:sec_ocr}

While financial communication heavily relies on PDFs, existing benchmarks \cite{xie2024finben} focus on visual QA or chart reasoning rather than document-structured text recovery.
We introduce the first multilingual financial OCR task, targeting end-to-end extraction of structured content from scanned filings, regulatory documents, and financial white papers.
This task bridges the gap between visual understanding and textual reasoning, enabling evaluation of multimodal systems on real-world document pipelines.

\paragraph{Task Definition.}  
The OCR task is defined as a structured information extraction problem from document images. Each financial PDF document is segmented into a set of page-level images \( \{I_1, I_2, \ldots, I_n\} \), where each image \( I_i \) corresponds to a single page. The model processes each image individually and generates a corresponding HTML-formatted text sequence \( T_i \), such that \( T_i = \text{OCR}(I_i) \), preserving both textual content and document structure (e.g., headings, tables, and paragraphs). The goal is to evaluate structural fidelity and semantic recovery, not just text recognition.

\paragraph{Data Construction.}  
Following the task design, we construct four novel datasets in a multilingual setting: \textit{EnglishOCR}, \textit{JapaneseOCR}, \textit{SpanishOCR}, and \textit{GreekOCR} (Table~\ref{tab:tab_datasets}).
The \textit{EnglishOCR} dataset is built using U.S. SEC EDGAR filings\footnote{\url{https://www.sec.gov/search-filings}}, which are primarily distributed in HTML format. For documents without native PDF versions, we use \texttt{wkhtmltopdf} to render PDF files from the corresponding HTML sources. Each PDF is segmented into page-level PNG images aligned with its  corresponding HTML content. To match each image with the most relevant HTML snippet, we compute cosine similarity between OCR-extracted text and HTML sentences using a Sentence-BERT (SBERT) model, resulting in 7,961 aligned image–HTML pairs.
The \textit{JapaneseOCR}, \textit{SpanishOCR}, and \textit{GreekOCR} datasets are constructed using source PDFs from Japanese Financial Services Agency (FSA) white papers\footnote{\url{https://www.fsa.go.jp/en/}}, Peruvian public regulatory documents\footnote{\url{https://www.bvl.com.pe/en/home-general}}, and Athens Stock Exchange annual company filings\footnote{\url{https://www.athexgroup.gr/en/market-data/issuers}}, respectively. Each PDF is decomposed into page-level PNG images, and the corresponding HTML content is generated by applying OCR to extract content and wrapping it in HTML tags to preserve document structure. We further apply quality-control by removing pages with excessive OCR errors and post-processing malformed outputs, such as repeated or garbled characters. After filtering, these datasets contain 17,586, 12,819, and 6,533 aligned image–HTML pairs, respectively. Detailed statistics are provided in Table~\ref{tab:tab_ocr} and Appendix~\ref{sec:sec_ocr_statistics}.

\paragraph{Evaluation Metric.}
% We employ ROUGE-1~\cite{lin-2004-rouge} to measure lexical and structural overlap between predicted and reference HTML sequences, providing a transparent metric for multilingual OCR quality. 
We employ ROUGE-1~\cite{lin-2004-rouge} to measure lexical and structural overlap between predicted and reference HTML sequences, complemented by human evaluation (Appendix~\ref{sec:sec_ocr_human_evaluation}) for reliable factuality assessment.

\section{Experimental Results}
\label{sec:Experiments}
To rigorously evaluate the effectiveness of \textsc{\multifinben} in exposing model limitations and guiding future development, we structure our experiments around four key research questions (RQs): 
%This approach not only reports raw performance but also provides defensive analyses, ablations, and insights to underscore \textsc{\multifinben}'s role as a diagnostic tool for multilingual and multimodal financial AI.
\textbf{RQ1:} How severe are the performance deficits of current LLMs when confronted with the multilingual and multimodal scenarios ubiquitous in real-world finance?
\textbf{RQ2:} Are there significant biases or trade-offs in model performance across different modalities?
\textbf{RQ3:} Can the model’s monolingual capabilities be effectively generalized to complex multilingual tasks that require cross-lingual comprehensive reasoning?
\textbf{RQ4:} How does our difficulty-aware framework help analyze the strengths and weaknesses of the model in more detail?

\paragraph{Evaluation Models.}
We evaluate 21 models (Table~\ref{tab:tab_models}, Appendix~\ref{sec:sec_models}) spanning text, vision, audio, and multimodal modalities, considering their openness under the Model Openness Framework (MOF, Appendix~\ref{sec:mof}) \cite{mof2024white}. %\cn{Textual tasks (see app1), Vision/OCR  (see app2), Audio  (see app3) xxx.  This diverse set is to ensure comprehensive coverage and highlights \textsc{\multifinben}'s ability to differentiate across openness levels.}
% For comparison, we include closed-source models, GPT-4o \cite{hurst2024gpt} and o3-mini \cite{openaio3mini}, which do not meet MOF Class III. The open-source models mostly fall under Class III, including multilingual multimodal models (Llama-4 \cite{llama4meta}, Gemma-3 series \cite{gemmateam2025gemma3technicalreport}, Qwen2.5-Omni \cite{Qwen2.5-Omni}), text-only multilingual models (Llama-3.1-70B \cite{dubey2024llama}, Deepseek-V3 \cite{liu2024deepseek}, Qwen2.5 \cite{qwen2.5}), and monolingual financial models (FinMA \cite{xie2023pixiu}, XuanYuan \cite{xuanyuan2024}, DeepSeek-R1-Distill-Japanese \cite{cyberagent-deepseek-r1-distill-qwen-32b-japanese}, FinMA-ES \cite{10.1145/3637528.3671554}, Plutus \cite{peng2025plutus}). For vision-language, we assess Qwen-VL-Max \cite{Qwen-VL}, DeepSeek-VL \cite{lu2024deepseekvl}, and LLaVA-v1.6-Vicuna-13B \cite{liu2023improved}, the only model partially meeting Class II. Audio-language models include Whisper-V3 \cite{pmlr-v202-radford23a}, Qwen2-Audio-7B, Qwen2-Audio-7B-Instruct \cite{chu2024qwen2audiotechnicalreport}, and SALMONN-7B/13B \cite{tang2024salmonngenerichearingabilities}.

\paragraph{Implementation Details.}
\label{sec:Implementation_details}

To ensure evaluation integrity and consistency, we customize our evaluation pipeline based on the LM Evaluation Harness \cite{eval-harness}.
OpenAI and TogetherAI-hosted models (DeepSeek, Llama-4, and Gemma-3) are accessed via their official APIs with temperature set to 0.
All other open-source models are deployed and evaluated locally using vLLM~\cite{kwon2023efficient} on GPUs.
Including the OpenAI and TogetherAI API costs, the total expenditure amounts to approximately \$80,000.

\paragraph{Key Results.} Table~\ref{tab:tab_results} presents model performance on the \textsc{\multifinben} benchmark\footnote{Results are also visualized on our leaderboard. For more details, refer to Appendix~\ref{sec:sec_leaderboard}.}. To ensure fair comparison, we additionally report a modality-balanced overall score for each model. The major findings are summarized as follows.

\begin{table*}[!htbp]
\centering
\small
\setlength{\textfloatsep}{5pt}
\renewcommand{\arraystretch}{0.95}

\begin{threeparttable}
\begin{adjustbox}{width=\textwidth}
\begin{tabular}{@{}cccc rrrrrrrrrrrrrrrrrrrrr@{}}
\toprule
\rotatebox{80}{\textbf{Modality}} & \rotatebox{80}{\textbf{Language}} & \rotatebox{80}{\textbf{Task}} & \rotatebox{80}{\textbf{Dataset}} &
\rotatebox{80}{\textbf{GPT-4o}} & \rotatebox{80}{\textbf{o3-mini}} & \rotatebox{80}{\textbf{Deepseek-V3}} &
\rotatebox{80}{\textbf{Llama-4-Scout}} & \rotatebox{80}{\textbf{Llama-3.1-70B}} & \rotatebox{80}{\textbf{Gemma-4B}} & \rotatebox{80}{\textbf{Gemma-27B}} &
\rotatebox{80}{\textbf{Qwen2.5-32B}} & \rotatebox{80}{\textbf{Qwen2.5-Omni}} &
\rotatebox{80}{\textbf{FinMA-7B}} &
\rotatebox{80}{\textbf{XuanYuan}} & \rotatebox{80}{\textbf{R1-Qwen32B-JA}} &
\rotatebox{80}{\textbf{FinMA-ES}} & \rotatebox{80}{\textbf{Plutus-8B}} &
\rotatebox{80}{\textbf{LLaVA-1.6}} & \rotatebox{80}{\textbf{Deepseek-VL}} &
\rotatebox{80}{\textbf{Whisper-V3}} & \rotatebox{80}{\textbf{Qwen2-Audio}} & \rotatebox{80}{\textbf{Qwen2-Audio-Inst}} &
\rotatebox{80}{\textbf{SALMONN-7B}} & \rotatebox{80}{\textbf{SALMONN-13B}} \\
\hline

\multirow{36}{*}{Text} & \multirow{13}{*}{EN} & IE & SC & \highlight{88.00} & 0.00 & 0.00 & 20.73 & \highlight{87.00} & 0.69 & 0.00 & 22.28 & 18.61 & \highlight{56.62} & 24.09 & 15.92 & 52.70 & 19.79 & 0.00 & 0.00 & 0.00 & 0.00 & 0.00 & 0.00 & 0.00\\
~ & ~ & IE & FinRED & \highlight{3.00} & 0.00 & 0.00 & 0.00 & 0.00 & 0.00 & 0.00 & \highlight{0.37} & 0.09 & 0.00 & 0.00 & 0.00 & 0.00 & \highlight{0.75} &0.00 & 0.00 & 0.00 & 0.00 & 0.00 & 0.00 & 0.00\\
~ & ~ & IE & FINER-ORD & \highlight{78.00} & 9.58 & 0.18 & 2.23 & \highlight{18.00} & 0.00 & 0.00 & \highlight{28.30} & 8.30 & 0.04 & 0.00 & 0.00 & 0.00 & 5.35 &0.00 & 0.00 & 0.00 & 0.00 & 0.00 & 0.00 & 0.00\\
~ & ~ & TA & Headlines & 78.00 & 0.00 & 47.32 & 84.33 & 60.00 & 0.00 & 0.00 & \highlight{85.42} & 82.23 & \highlight{97.08} & 85.10 & 82.21 & \highlight{94.69} & 71.14 &0.00 & 0.00 & 0.00 & 0.00 & 0.00 & 0.00 & 0.00\\
~ & ~ & TA & TSA &61.00 & 0.00 & 0.85 & 63.82 & 42.00 & 32.34 & 32.34 & 42.98 & 43.40 & \highlight{81.70} & \highlight{85.11} & 60.00 & \highlight{86.38} & 54.89 &0.00 & 0.00 & 0.00 & 0.00 & 0.00 & 0.00 & 0.00\\
~ & ~ & QA & XBRL-Math & \highlight{68.00} & \highlight{68.89} & \highlight{67.78} & 27.78 & 62.00 & 0.00 & 11.11 & 64.44 & 44.44 & 7.78 & 3.33 & 2.22 & 6.67 & 14.44 &0.00 & 0.00 & 0.00 & 0.00 & 0.00 & 0.00 & 0.00\\
~ & ~ & QA & FinQA & \highlight{5.00}& 0.00& 0.00& 0.00& 0.00 & 0.00& 0.00& 0.00& 0.00& \highlight{7.41}& 0.00& 0.00& 0.00& \highlight{1.22}&0.00 & 0.00 & 0.00 & 0.00 & 0.00 & 0.00 & 0.00\\
~ & ~ & QA & TATQA & 0.00 & 0.00& 0.00& 0.36& \highlight{44.00}& 0.00&0.00 &0.05 &1.73 &\highlight{4.14} &0.00 &0.00 &0.00 &\highlight{15.16} &0.00 & 0.00 & 0.00 & 0.00 & 0.00 & 0.00 & 0.00\\
~ & ~ & TG & ECTSUM & 0.00&0.00 &0.00 &0.00 &0.00 &0.00 &0.00 &0.00 &0.00 &0.00 &0.00 &0.00 &0.00 &0.00 &0.00 & 0.00 & 0.00 & 0.00 & 0.00 & 0.00 & 0.00\\
~ & ~ & TG & EDTSUM & \highlight{25.00} & 19.13& 16.80&16.59 &18.00 &0.98 &0.10 &\highlight{20.16} &\highlight{23.89} &19.92 &12.49 &8.06 &2.06 &13.61 &0.00 & 0.00 & 0.00 & 0.00 & 0.00 & 0.00 & 0.00\\
~ & ~ & RM & CCF & \highlight{52.50} & 50.00 & 50.62 & \highlight{51.34} & 50.00 & 50.93 & 50.00 & \highlight{52.94} & 50.31 & 50.05 & 50.00 & 50.00 & 51.18 & 50.00 &0.00 & 0.00 & 0.00 & 0.00 & 0.00 & 0.00 & 0.00\\
~ & ~ & PO & BigData22 & 48.50 & 50.00 & 50.93 & 46.91 & 50.00 & 50.75 & 50.00 & 49.89 & \highlight{51.82} & 50.80 & \highlight{53.12} & 50.00 & \highlight{52.12} & 50.26 &0.00 & 0.00 & 0.00 & 0.00 & 0.00 & 0.00 & 0.00\\
~ & ~ & DM & MSFT & 41.32 & 65.06 & 0.00 & 0.00 & 72.25 & \highlight{74.03} & \highlight{79.97} & 49.32 & 0.00 & 0.00 & 68.81 & \highlight{74.50} & 66.53 & 65.10 & 0.00 & 0.00 & 0.00 & 0.00 & 0.00 & 0.00 & 0.00\\

% \cmidrule(lr){3-25}
% ~ & ~ & ~ & Average & \highlight{42.18} & 20.20 & 18.04 & 24.16 & \highlight{38.71} & 16.13 & 17.19 & \highlight{32.01} & 24.99 & 28.89 & 29.39 & 26.38 & 31.72 & 27.82 & 0.00 & 0.00 & 0.00 & 0.00 & 0.00 & 0.00 & 0.00 \\

\cmidrule(lr){2-25}
~ & \multirow{4}{*}{ZH} & IE & RRE  &\highlight{63.25} &0.00 &\highlight{67.52} &\highlight{54.70} &46.15 &36.75 &36.75 &8.55 &7.69 &0.85 &2.56 &0.85 &0.85 &2.56 &0.00 & 0.00 & 0.00 & 0.00 & 0.00 & 0.00 & 0.00\\
~ & ~ & TA & AIE  & \highlight{82.26}&0.00 &\highlight{82.01} &80.99 &76.80 &33.82 &33.82 &\highlight{83.03} &80.17 &40.81 &10.04 &4.32 &21.55 &54.48 &0.00 & 0.00 & 0.00 & 0.00 & 0.00 & 0.00 & 0.00\\
~ & ~ & TA & LNE  & \highlight{63.30}& 0.00& \highlight{58.72}&55.50 &41.28 &9.17 & 9.17&57.80 &\highlight{59.17} &29.82 &22.48 &12.84 &32.11 &26.61 & 0.00 & 0.00 & 0.00 & 0.00 & 0.00 & 0.00 & 0.00\\
~ & ~ & QA & FinancialIQ  &32.53&0.00 &35.52 &\highlight{66.83} &62.71 &25.19 &25.20 &\highlight{77.09} & \highlight{65.32} &26.21 & 57.07&34.70 &31.48 &40.52 &0.00 & 0.00 & 0.00 & 0.00 & 0.00 & 0.00 & 0.00\\

% \cmidrule(lr){3-25}
% ~ & ~ & ~ & Average & \highlight{60.34} & 0.00 & \highlight{60.94} & \highlight{64.51} & 56.74 & 26.23 & 26.24 & 56.62 & 53.09 & 24.42 & 23.04 & 13.18 & 21.50 & 31.04 & 0.00 & 0.00 & 0.00 & 0.00 & 0.00 & 0.00 & 0.00 \\

\cmidrule(lr){2-25}
~ & JA & TA & chabsa  &0.00 &0.00 &0.00 &\highlight{48.43} &32.17 &8.98 &23.96 &4.54 &44.35 &46.94&\highlight{47.59} &23.96 &\highlight{57.36} &34.62 &0.00 & 0.00 & 0.00 & 0.00 & 0.00 & 0.00 & 0.00\\

\cmidrule(lr){2-25}
~ & \multirow{4}{*}{ES} & TA & MultiFin  & \highlight{61.74}&0.00 &53.91 &\highlight{62.17} &48.26 &22.17 &22.17 & 46.52&46.96 &43.04 &31.74 &12.61 &44.78 &\highlight{51.30} &0.00 & 0.00 & 0.00 & 0.00 & 0.00 & 0.00 & 0.00\\
~ & ~ & TA & TSA  & 0.39&0.00 &29.17 &52.29 &24.29 &\highlight{63.04} &\highlight{63.46} &31.63 &46.46 &31.03 &\highlight{68.19} & 63.38&16.64 &51.82 &0.00 & 0.00 & 0.00 & 0.00 & 0.00 & 0.00 & 0.00\\
~ & ~ & QA & EFPA  & 31.14 & 0.00 & 18.86 & \highlight{67.54} & \highlight{66.67} & 25.44 & 25.44 & 65.79 & 55.70 & 32.46 & 65.79 & 25.44 & \highlight{91.67} & 48.25 &0.00 & 0.00 & 0.00 & 0.00 & 0.00 & 0.00 & 0.00\\
~ & ~ & TG & FNS-2023  & \highlight{25.94} & \highlight{18.11} & 0.00 & 9.61 & \highlight{12.14} & 0.00 & 0.00 & 5.93 & 7.50 & 1.64 & 5.71 & 10.62 & 1.65 & 9.27 &0.00 & 0.00 & 0.00 & 0.00 & 0.00 & 0.00 & 0.00 \\

% \cmidrule(lr){3-25}
% ~ & ~ & ~ & Average & 29.80 & 4.53 & 25.49 & \highlight{47.90} & 37.84 & 27.66 & 27.77 & 37.47 & 39.16 & 27.04 & \highlight{42.86} & 28.01 & 38.69 & \highlight{40.16} & 0.00 & 0.00 & 0.00 & 0.00 & 0.00 & 0.00 & 0.00 \\

\cmidrule(lr){2-25}
~ & \multirow{4}{*}{EL} & IE & GRFinNUM  & 9.18 & 20.98 & 7.43 & \highlight{49.12} & \highlight{46.34} & 0.00 & 0.00 & \highlight{36.77} & 0.40 & 0.00 & 0.00 & 0.00 & 0.00 & 70.06 & 0.00 & 0.00 & 0.00 & 0.00 & 0.00 & 0.00 & 0.00\\
~ & ~ & QA & GRFinQA  & \highlight{78.22} & 0.00 & 50.00 & \highlight{74.22} & \highlight{64.44} & 22.67 & 22.67 & 60.44 & 48.89 & 25.33 & 57.78 & 28.44 & 23.11 & 64.00 & 0.00 & 0.00 & 0.00 & 0.00 & 0.00 & 0.00 & 0.00\\
~ & ~ & TG & GRFNS-2023  & \highlight{25.50} & 16.95 & \highlight{37.72} & 16.90 & 13.61 & 0.24 & 0.21 & 9.71 & 5.60 & 11.20 & 6.48 & 14.45 & 3.56 & \highlight{34.46} & 0.00 & 0.00 & 0.00 & 0.00 & 0.00 & 0.00 & 0.00\\
~ & ~ & TA & GRMultiFin  & 59.26 & 0.00 & \highlight{61.11} & 55.56 & 50.00 & 38.89 & 38.89 & \highlight{70.37} & 38.89 & 35.19 & 53.70 & 40.74 & 35.19 & \highlight{72.22} & 0.00 & 0.00 & 0.00 & 0.00 & 0.00 & 0.00 & 0.00\\

% \cmidrule(lr){3-25}
% ~ & ~ & ~ & Average & 43.04 & 9.48 & 39.07 & \highlight{48.95} & 43.60 & 15.45 & 15.44 & \highlight{44.32} & 23.45 & 17.93 & 29.49 & 20.91 & 15.47 & \highlight{60.19} & 0.00 & 0.00 & 0.00 & 0.00 & 0.00 & 0.00 & 0.00 \\

\cmidrule(lr){2-25}
~ & BI & TG & DOLFIN  &\highlight{92.29} & 90.13 & 86.26 & 89.17 & \highlight{92.13} & 35.92 & 35.92 & \highlight{92.29} & 91.80 & 69.24 & 91.60 & 71.81 & 66.57 & 91.59 & 0.00 & 0.00 & 0.00 & 0.00 & 0.00 & 0.00 & 0.00\\

\cmidrule(lr){2-25}
~ & \multirow{2}{*}{MU} & QA & PolyFiQA-Easy 
& 9.79 & 9.56 & \highlight{34.72} & \highlight{27.73} & \highlight{25.04} 
& 15.02 & 14.74 & 19.34 & 18.81 & 2.44 & 2.40 & 11.63 & 0.63 & 7.06 
& 0.00 & 0.00 & 0.00 & 0.00 & 0.00 & 0.00 & 0.00\\

~ & ~ & QA & PolyFiQA-Expert 
& 5.31 & 4.85 & \highlight{30.35} & \highlight{20.60} & \highlight{18.56} 
& 13.83 & 16.01 & 18.17 & 16.35 & 6.38 & 0.71 & 8.80 & 0.00 & 9.87 
& 0.00 & 0.00 & 0.00 & 0.00 & 0.00 & 0.00 & 0.00\\

% \cmidrule(lr){3-25}
% ~ & ~ & ~ & Average 
% & 7.55 & 7.21 & \highlight{32.54} & \highlight{24.17} & \highlight{21.80} 
% & 14.43 & 15.38 & 18.76 & 17.58 & 4.41 & 1.56 & 10.22 & 0.32 & 8.47 
% & 0.00 & 0.00 & 0.00 & 0.00 & 0.00 & 0.00 & 0.00 \\

\cmidrule(lr){2-25}
~ & ~ & ~ & Average 
& \highlight{40.98} & 14.59 & 30.61 & \highlight{39.50} & \highlight{42.20} & 19.34 & 20.41 & 38.07 
& 33.06 & 26.83 & 31.24 & 24.40 & 28.95 & 35.53 
& 0.00 & 0.00 & 0.00 & 0.00 & 0.00 & 0.00 & 0.00 \\

\midrule
\multirow{8}{*}{Vision} 
& \multirow{2}{*}{EN} & IE & EnglishOCR  
& \highlight{21.38} & 0.00 & 0.00 & \highlight{12.39} & 0.00 & \highlight{10.70} & 11.40 & 0.00 & 0.00 & 0.00 & 0.00 & 0.00 & 0.00 & 0.00 & 0.00 & 0.00 & 0.00 & 0.00 & 0.00 & 0.00 & 0.00\\

~ & ~ & QA & TableBench  
& \highlight{66.70} & 0.00 & 0.00 & 32.30 & 0.00 & 28.60 & 60.90 & 0.00 & \highlight{74.90} & 0.00 & 0.00 & 0.00 & 0.00 & 0.00 & \highlight{59.30} & 57.30 & 0.00 & 0.00 & 0.00 & 0.00 & 0.00\\

\cmidrule(lr){2-25}
~ & JA & IE & JapaneseOCR  
& 21.63 & 0.00 & 0.00 & \highlight{24.52} & 0.00 & \highlight{25.82} & \highlight{26.72} & 0.00 & 21.59 & 0.00 & 0.00 & 0.00 & 9.70 & 6.62 & 0.00 & 0.00 & 0.00 & 0.00 & 0.00 & 0.00 & 0.00\\

\cmidrule(lr){2-25}
~ & ES & IE & SpanishOCR  
& \highlight{78.55} & 0.00 & 0.00 & 4.12 & 0.00 & \highlight{5.60} & \highlight{4.40} & 0.00 & 0.00 & 0.00 & 0.00 & 0.00 & 0.00 & 0.00 & 0.00 & 0.00 & 0.00 & 0.00 & 0.00 & 0.00 & 0.00\\

\cmidrule(lr){2-25}
~ & EL & IE & GreekOCR  
& \highlight{41.86} & 0.00 & 0.00 & \highlight{42.51} & 0.00 & 23.69 & \highlight{30.60} & 0.00 & 0.00 & 0.00 & 0.00 & 0.00 & 8.25 & 6.80 & 0.00 & 0.00 & 0.00 & 0.00 & 0.00 & 0.00 & 0.00\\

\cmidrule(lr){2-25}
~ & ~ & ~ & Average 
& \highlight{46.02} & 0.00 & 0.00 & \highlight{23.17} & 0.00 & 18.88 & \highlight{26.80} & 0.00 & 19.30 & 0.00 & 0.00 & 0.00 & 0.00 & 0.00 & 15.45 & 14.14 & 0.00 & 0.00 & 0.00 & 0.00 & 0.00\\

\midrule
\multirow{3}{*}{Audio} & \multirow{2}{*}{EN} & TG & MDRM-test  & 95.77 & 0.00 & 0.00 & 0.00 & 0.00 & 0.00 & 0.00 & 0.00 & \highlight{96.43} & 0.00 & 0.00 & 0.00 & 0.00 & 0.00 & 0.00 & 0.00 & \highlight{97.86} & \highlight{96.03} & 95.32 & 48.48 & 49.17\\
~ & ~ & TG & FinAudioSum & \highlight{6.30} & 0.00 & 0.00 & 0.00 & 0.00 & 0.00 & 0.00 & 0.00 & 0.00 & 0.00 & 0.00 & 0.00 & 0.00 & 0.00 & 0.00 & 0.00 & \highlight{5.30} & 0.00 & \highlight{4.80} & 0.00 & 0.00\\

\cmidrule(lr){2-25}
~ & ~ & ~ & Average & \highlight{51.04} & 0.00 & 0.00 & 0.00 & 0.00 & 0.00 & 0.00 & 0.00 & \highlight{48.22} & 0.00 & 0.00 & 0.00 & 0.00 & 0.00 & 0.00 & 0.00 & \highlight{51.58} & 48.02 & 50.06 & 24.24 & 24.59\\

\midrule
~ & ~ & ~ & Modality-Balanced Average 
& \highlight{46.01} & 4.86 & 10.20 & \highlight{20.89} & 14.07 & 12.74 & 15.74 & 12.69 
& \highlight{33.53} & 8.94 & 10.41 & 8.13 & 9.65 & 11.84 
& 5.15 & 4.71 & 17.19 & 16.01 & 16.69 & 8.08 & 8.20 \\

\bottomrule
\end{tabular}
\end{adjustbox}

% \vspace{0.5em}
\end{threeparttable}
\caption{
Standardized performance of evaluated LLMs on the \textsc{\multifinben}.
The final row presents the modality-balanced average, computed as the mean of text, vision, and audio averages.
}
\label{tab:tab_results}
\end{table*}

\begin{figure*}[ht]
    \centering
    % Subfigure 1 - Modality
    \begin{minipage}{0.32\linewidth}
        \includegraphics[width=\linewidth]{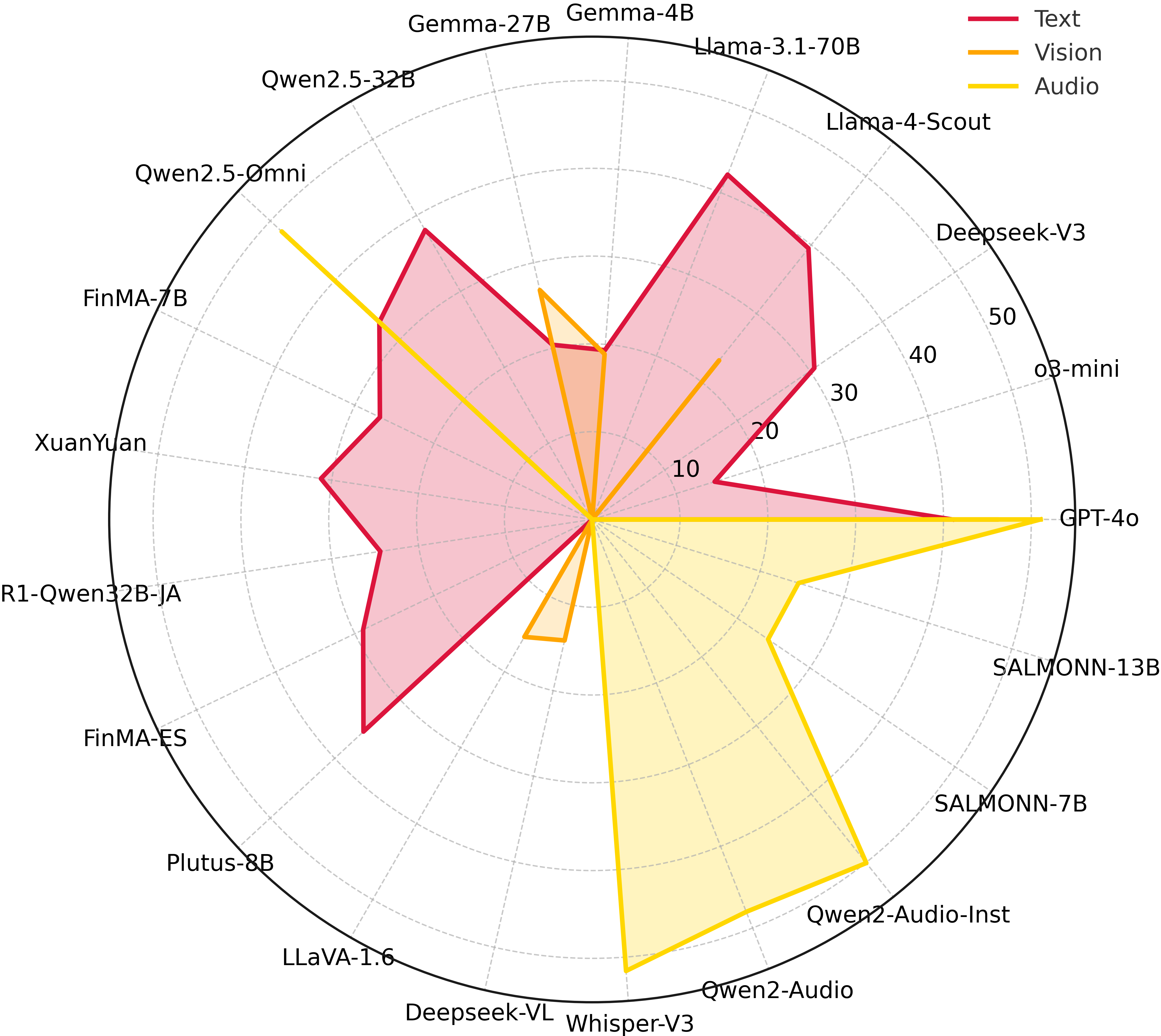}
        \caption{\small Performance across modalities: Text, Vision, Audio.}
        \label{fig:modality_radar}
    \end{minipage}
    % Subfigure 2 - Language
    \begin{minipage}{0.32\linewidth}
        \includegraphics[width=\linewidth]{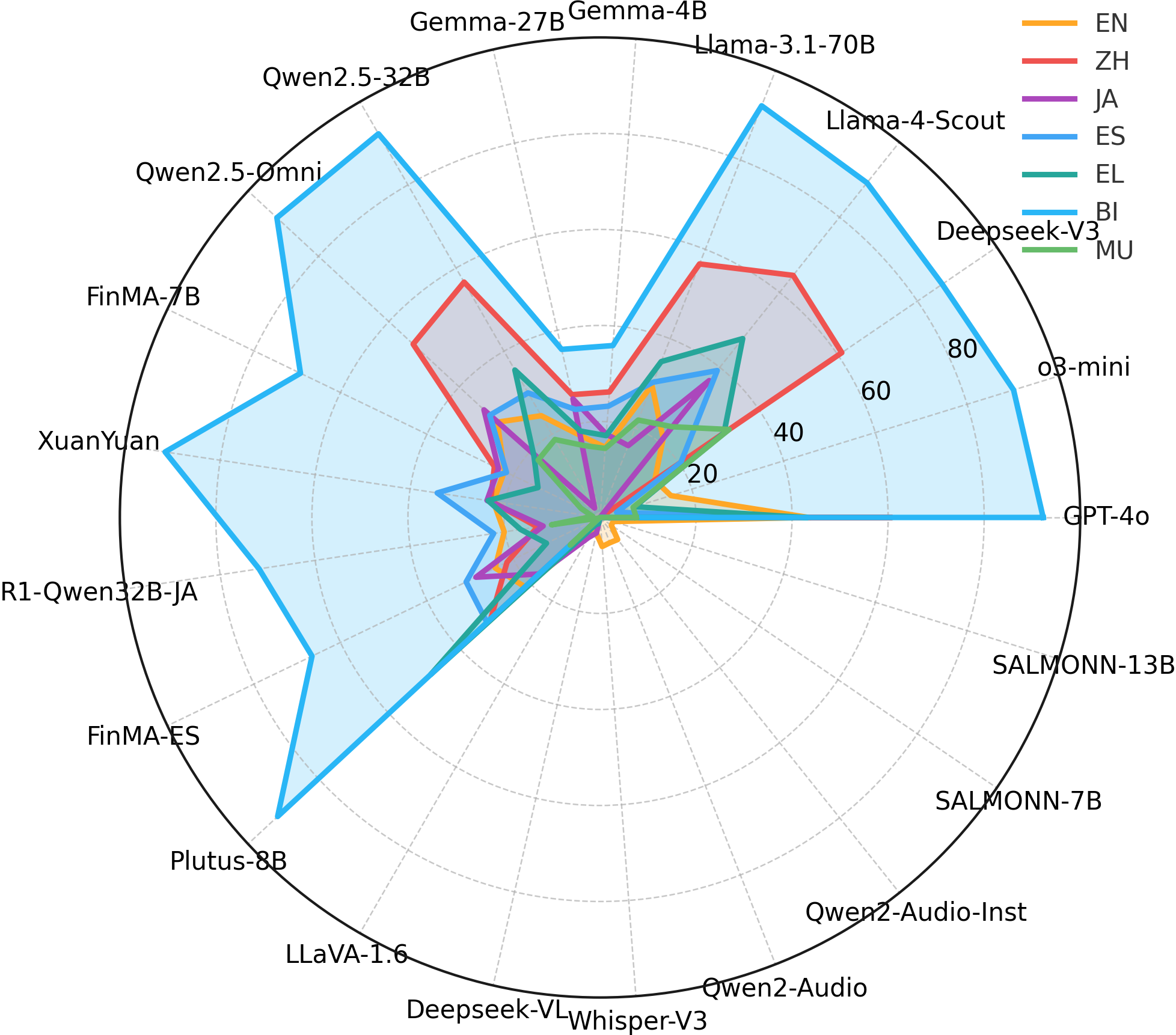}
        \caption{\small Performance across languages: EN, ZH, JA, ES, EL, BI, MU.}
        \label{fig:language_radar}
    \end{minipage}
    % Subfigure 3 - Difficulty
    \begin{minipage}{0.32\linewidth}
        \includegraphics[width=\linewidth]{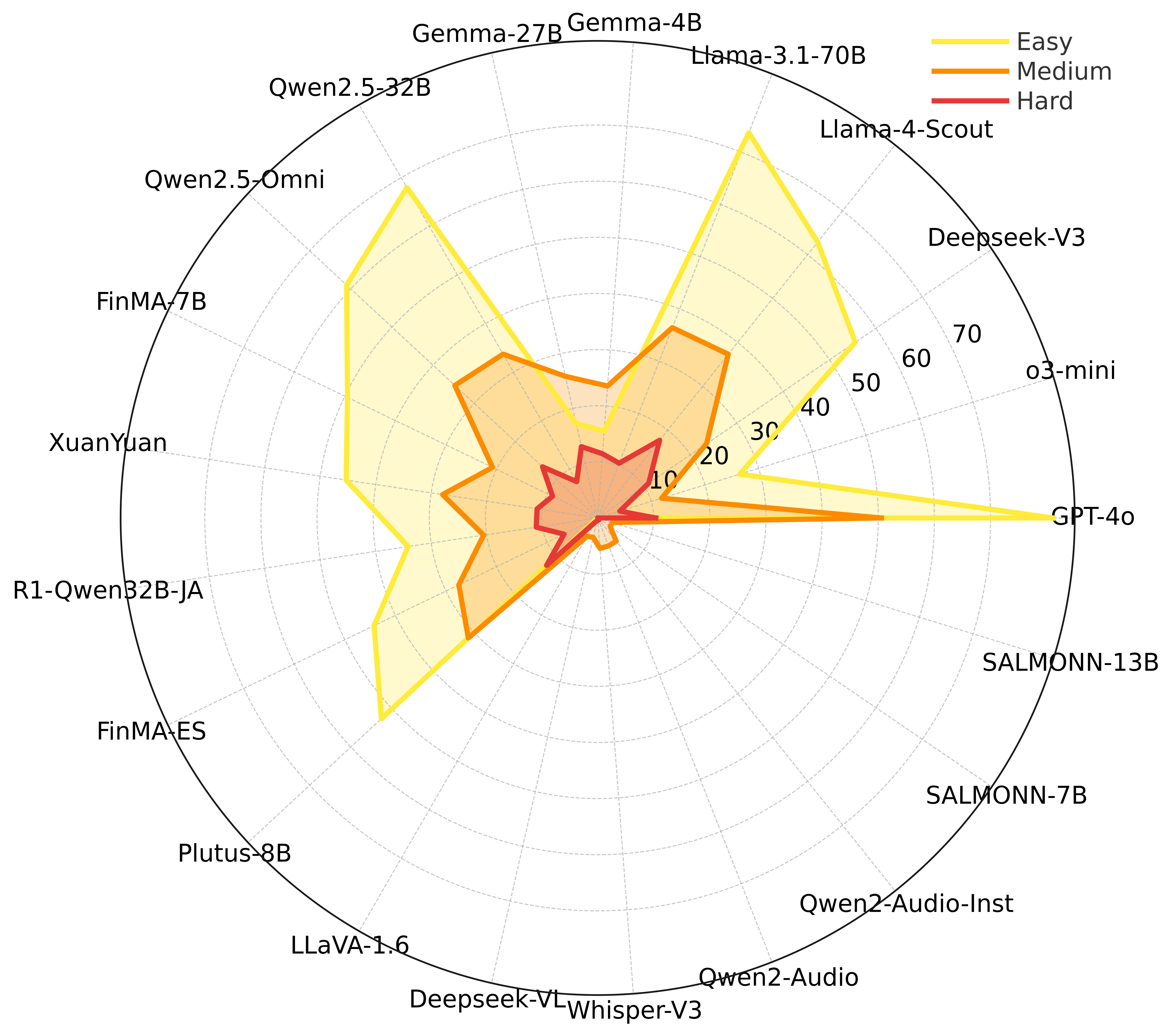}
        \caption{\small Performance across difficulty levels: Easy, Medium, Hard.}
        \label{fig:difficulty_radar}
    \end{minipage}

    \caption{Radar charts comparing model performance across (a) modalities, (b) languages, and (c) difficulty levels. The figures demonstrate diverse strengths and limitations of models in various dimensions.}
    \label{fig:all_radar}
\end{figure*}
\emph{Finding 1 for RQ1: Cross-modal and cross-lingual reasoning in \textsc{\multifinben} exposes clear limits of current LLMs.}
Building upon our structured, difficulty-aware benchmarking framework, \textsc{\multifinben} consists of 6 easy, 18 medium, and 12 hard datasets spanning three modalities and five economically critical languages across three linguistic settings. All multilingual, vision, and audio datasets fall into the medium or hard tiers, posing nontrivial challenges even for the most advanced models. The leading model, GPT-4o, achieves an overall score of only 46.01\%, revealing clear limitations even at the frontier of multimodal and multilingual reasoning. The next best models, Qwen2.5-Omni and Llama-4, reach 33.53\% and 20.89\%, respectively, underscoring the steep difficulty gradient established by our benchmark. Notably, all three top-performing systems are multimodal and multilingual, reinforcing the necessity of these capabilities for robust financial reasoning.
In contrast, monomodal and monolingual models perform considerably worse. The strongest text-only model, Llama-3.1-70B, attains merely 14.07\% overall (with a best text-modality score of 42.20\%). Likewise, modality- or language-specific models such as Whisper-V3 (audio-only, 17.19\%), LLaVA-1.6 (vision-only, 5.15\%), and FinMA-7B (English-only, 8.94\%) trail far behind. These disparities highlight the inherent limitations of models lacking cross-modal and cross-lingual reasoning, emphasizing the critical need for integrated multimodal and multilingual understanding in financial AI, precisely the kind of comprehensive evaluation that \textsc{\multifinben} is designed to provide.

\emph{Finding 2 for RQ2: Multimodal models exhibit trade-offs across modalities.}
Figure~\ref{fig:modality_radar} illustrates the stratified performance across modalities. 
While GPT-4o leads overall performance on \textsc{\multifinben}, it is surpassed by the text-specialized Llama-3.1-70B in text-only tasks (42.20\% vs. 40.98\%). Similarly, other leading multimodal models such as Qwen2.5-Omni and Llama-4 fall short of their text-only counterparts Qwen2.5-32B (38.07\% vs. 33.06\%) and Llama-3.1-70B (42.20\%  vs. 39.50\%) in textual evaluations. This trade-off becomes more pronounced in mid-tier models like Gemma-3-4B and Gemma-3-27B, which show comparable vision-task performance (18.88\% and 26.80\%) to vision-only models (LLaVA-1.6: 15.45\%, Deepseek-VL: 14.14\%) but substantially lower text-task performance (12.74\% and 15.74\%). These trends highlight the difficulty of preserving text-specific optimization when expanding to multimodal capabilities.
In contrast, multimodal models decisively outperform unimodal baselines in vision and audio tasks, i.e, GPT-4o achieves 46.02\% in vision and 51.04\% in audio, surpassing modality-specific counterparts. The asymmetric performance reveals that while text tasks benefit from data maturity, vision and audio gain more from multimodal integration, making unified models essential for complex financial real-world applications.

\emph{Finding 3 for RQ3: Scaling monolingual ability does not yield multilingual reasoning.}
Figure~\ref{fig:language_radar} illustrates the stratified performance across linguistic settings.
Although GPT-4o consistently ranks among the top performers across individual monolingual settings, it performs poorly on multilingual tasks (9.79\% on \textit{PolyFiQA-Easy} and 5.31\% on \textit{PolyFiQA-Expert}; see detailed error analysis in Appendix~\ref{sec:sec_error_analysis}). A similar pattern holds for Qwen2.5-Omni (18.81\% and 16.35\%), suggesting that strong monolingual proficiency does not necessarily translate into multilingual reasoning.
In contrast, models with lower overall benchmark performance, such as Deepseek-V3 (34.72\% and 30.35\%) and Llama-3.1-70B (25.04\% and 18.56\%), achieve markedly better results on multilingual tasks.
These findings indicate that multilingual reasoning constitutes a fundamentally different challenge from monolingual evaluation, one that cannot be addressed simply by scaling monolingual capacity but instead requires dedicated modeling for multilingual understanding and reasoning.
Given the increasing demand for simultaneous multilingual comprehension in global financial contexts, \textsc{\multifinben}, particularly through its multilingual component (\textit{PolyFiQA-Easy} and \textit{PolyFiQA-Expert}, the first multilingual financial datasets), provides a critical stress test for this capability, capturing dimensions of multilingual reasoning that isolated monolingual evaluations fail to reveal.

% \paragraph{Monolingual models underperform in multilingual settings.}
% English-only models, such as finma-7b-full (8.91\%, 15nd), consistently trail bilingual/multilingual models like Llama3.1-XuanYuan (10.42\%, 12th), FinMA-ES (9.65\%, 14th), and plutus (11.82\%, 10th). This performance gap widens between FinMA-7b-full and plutus further in bilingual (69.24\% vs. 91.59\%) and multilingual (3.10\% vs. 7.24\%) evaluation scenarios, reflecting the critical impact of cross-lingual capacity. Moreover, language-specific performance aligns closely with training data coverage. High-resource languages (e.g., English, Chinese, Japanese, Spanish) are dominated by generalist frontier models such as GPT-4o and Llama-4, while domain-specialized models like plutus achieve superior results in low-resource languages, as evidenced in Greek (60.19\% vs. Llama-4’s 48.95\%). These findings establish that multilinguality is not merely an auxiliary feature but a foundational requirement for LLMs in finance, positioning \textsc{\multifinben} as the first comprehensive benchmark to diagnose and drive progress in multilingual financial language understanding.

\emph{Finding 4 for RQ4: Difficulty-aware benchmarking enables stratified, fine-grained, and dynamic evaluation.}
Figure~\ref{fig:difficulty_radar} illustrates the stratified performance across difficulty levels.
\textsc{\multifinben} introduces a three-tier structure (easy, medium, and hard) enabling layered evaluation of model capabilities. Performance converges at the hard tier, where all models struggle, whereas the easy and medium tiers yield clearer distinctions in capability. Except for Gemma-3-4B and Gemma-3-27B, which outperform on medium datasets due to their multimodal strengths, most models follow the monotonic trend easy$>$medium$>$hard.
General-purpose models such as GPT-4o, Llama-4, and Qwen2.5-Omni demonstrate strong performance on easy and medium tiers but face substantial challenges on hard tasks. These hard-tier datasets (\textit{PolyFiQA-Easy}, \textit{PolyFiQA-Expert}, OCR-based, and other hard-tier datasets) require multilingual, multi-document reasoning and the joint interpretation of textual and structural information, posing a fundamental challenge for current models.
By contrast, domain-specific models such as o3-mini, XuanYuan, and R1-Qwen32B-JA underperform on easy datasets, suggesting that heavy specialization may compromise generalization, a consideration for future training paradigms.
Through this difficulty-aware design, \textsc{\multifinben} enables stratified and interpretable evaluation, revealing nuanced model strengths and weaknesses across complexity levels. As model capabilities advance, these tiers may shift dynamically, medium datasets may become easy, allowing the benchmark to evolve alongside model progress and real-world financial reasoning demands.

% \paragraph{Structured difficulty-aware benchmarking facilitates dynamic evaluation.}
% While early models such as finma-7b-full perform competitively on easy tasks (49.48\%), performance declines sharply on medium (22.01\%) and hard tasks (9.49\%), with newer models like GPT-4o showing improvements yet continuing to struggle. To address the lack of coverage in existing benchmarks, we introduce PolyFiQA-Easy and PolyFiQA-Expert, which are the first multilingual financial datasets, and the first financial OCR task, explicitly designed to target underrepresented linguistic and modality challenges. These datasets, initially created to fill specific gaps, emerged as some of the most difficult evaluation tasks, with Deepseek-V3 achieving only 42.58\% and 31.40\% on PolyFiQA-Easy and PolyFiQA-Expert, and overall model averages of 7.50\% and 5.61\%, respectively. This finding highlights how \textsc{\multifinben}’s structured design not only reveals current model limitations but also serves as a practical framework to systematically guide the development of new datasets and scenarios, ensuring benchmarks evolve in step with model capabilities and real-world financial demands.

\section{Conclusion}
We present \textsc{\multifinben}, the first multilingual and multimodal benchmark for evaluating real-world financial applications under difficulty-aware settings. Our results show persistent performance gaps across languages, modalities, and difficulty levels, with even GPT-4o exhibiting substantial limitations in complex scenarios. The newly introduced datasets (\textit{PolyFiQA} and \textit{OCR}) constitute the core contribution of \textsc{\multifinben} by addressing critical gaps in multilingual reasoning and document understanding from scanned financial reports. Together, they reveal weaknesses in cross-lingual and cross-modal reasoning. \textsc{\multifinben} provides a comprehensive foundation for developing inclusive, multilingual, and multimodal financial LLMs  capable of realistic expert-level reasoning.
% We introduce \textsc{\multifinben}, the first multilingual and multimodal benchmark designed to systematically evaluate real-world financial application under difficulty-aware settings. Our analysis reveals persistent performance gaps across modalities, languages, and difficulty levels, with even GPT-4o showing substantial limitations, particularly in complex multilingual and multimodal scenarios.
% Through our unified modality-language-task framework and the introduction of novel datasets, including PolyFiQA-Easy, PolyFiQA-Expert, EnglishOCR, JapaneseOCR, SpanishOCR, and GreekOCR, we stress-test existing models and expose nuanced limitations in long-form reasoning, audio understanding, OCR, and multilingual comprehension. The sharp performance degradation in multilingual settings demonstrates that scaling monolingual capacity alone is insufficient for global financial applications.
% \textsc{\multifinben} provides a comprehensive and dynamic evaluation platform and offers a foundation for future research toward inclusive, multilingual, and multimodal financial AI. By releasing all datasets and evaluation scripts, we aim to foster reproducible and impactful research and encourage the community to develop the next generation of financial LLMs capable of robust, cross-modal, and cross-lingual reasoning in real-world economic and policy contexts.

\section*{Limitations}
\label{sec:sec_limitations}

Despite the contributions of this study, several limitations remain:
\begin{itemize}
    \item \textbf{Limited Availability of High-Quality Open Datasets.} The financial domain suffers from a lack of truly open-source datasets. Even among publicly released datasets, many include ambiguous or restrictive licensing terms that hinder redistribution or standardized benchmarking. As a result, our current coverage is limited and may not fully represent the breadth of real-world financial tasks.
    \item \textbf{Relatively Small Scale of Newly Introduced Datasets.} While we introduce several novel datasets, their sample sizes are modest compared with large-scale benchmarks in general NLP. This constraint arises from the high annotation cost and domain expertise required for financial data. Future expansions with larger and more diverse samples will further enhance statistical robustness and task coverage.
    \item \textbf{Limited Language Coverage in the Audio Modality.} Our audio evaluation currently focuses on English earnings calls. Due to regional, copyright, and legal restrictions, obtaining and releasing multilingual earnings-call recordings remains challenging. Nevertheless, the U.S. market is globally representative, and our dataset spans three years, multiple sectors, and diverse downstream tasks, providing a meaningful testbed for financial speech understanding. Future extensions will incorporate multilingual earnings-call data to broaden cross-lingual evaluation of audio-based financial reasoning.
    \item \textbf{Incomplete Model Coverage.} Due to computational and access constraints, we evaluate only a curated subset of representative models. While these span diverse modalities, access types, and language capabilities, broader coverage of emerging models, particularly open-access and region-specific LLMs, would further strengthen future iterations of the benchmark.
    \item \textbf{No Baseline Adjustment During Metric Normalization.} In our difficulty-aware scoring and normalization process, we do not subtract performance from random or majority-class baselines. This may overstate absolute performance on certain tasks, especially those with class imbalance or low inherent complexity. Incorporating baseline-adjusted metrics in future releases would improve comparability across tasks.
\end{itemize}

\section*{Potential Risks}
\label{sec:sec_potential}

\paragraph{Potential Positive Societal Impacts.}
This work contributes a dynamic, multilingual, and multimodal benchmark for evaluating large language models in the financial domain. By enabling more rigorous and transparent comparison of models across diverse languages, modalities, and task types, this benchmark:
\begin{itemize}
    \item Promotes equitable access to financial AI by supporting underrepresented languages and non-English financial datasets.
    \item Encourages the development of safer, more reliable financial LLMs that can assist in decision-making, regulatory compliance, and financial literacy across global markets.
    \item Provides tools for better understanding model capabilities and limitations in high-stakes, real-world applications such as financial question answering, document analysis, and OCR.
\end{itemize}
By emphasizing open evaluation and releasing datasets and evaluation code, this work also fosters reproducibility and responsible innovation in financial AI.

\paragraph{Potential Negative Societal Impacts.}
Despite its contributions, this benchmark carries several potential risks:
\begin{itemize}
    \item Misuse or Overreliance: Improved performance on benchmark tasks may be mistaken for robust real-world competence. Overreliance on high-scoring models without thorough safety, fairness, or legal audits could lead to financial misinformation, biased decision-making, or consumer harm.
    \item Representation Bias: While we aim for multilingual coverage, some languages and financial contexts remain underrepresented. This may amplify biases or exclude important financial narratives from less-resourced regions.
    \item Data Privacy and Licensing Concerns: Although we carefully selected datasets for inclusion, ambiguities in licensing and potential data leakage from pretraining corpora may introduce ethical or legal risks if reused without careful scrutiny.
\end{itemize}
Ongoing community engagement and transparent documentation are essential to mitigate these concerns as the benchmark evolves.

\section*{Ethical Considerations and Usage Disclaimer}
\label{sec:sec_ethical}

The authors assume full responsibility for the development and dissemination of the MultiFinBen benchmark and any associated materials. All raw datasets included are publicly available and do not contain personal or sensitive information. We have made diligent efforts to ensure that the construction and release of MultiFinBen comply with established ethical standards and privacy guidelines.

All datasets provided as part of MultiFinBen are released under the MIT License, and users are expected to comply with the terms of that license.

This manuscript, along with any accompanying source code, datasets, and supplementary materials (collectively, the “Material”), is intended solely for academic and educational purposes. It is important to note that the Material does not constitute financial, legal, or investment advice, nor should it be relied upon for decision-making in any practical or commercial context.

While the authors have made reasonable efforts to ensure the accuracy and reliability of the Material, no express or implied warranties are provided regarding its completeness or suitability for any particular use. The authors and their affiliated institutions disclaim any liability for direct, indirect, incidental, or consequential damages that may result from the use or reliance on the Material. Users are strongly encouraged to seek professional advice when making financial, legal, or investment decisions.

By accessing or utilizing the Material, users agree to indemnify, defend, and hold harmless the authors and their affiliated organizations from any claims, damages, or liabilities arising from such use.

% \section*{Acknowledgments}

% This work has been partially supported by project MIS 5154714 of the National Recovery and Resilience Plan Greece 2.0 funded by the European Union under the NextGenerationEU Program. Shengyuan Colin Lin, Keyi Wang, and Xiao-Yang Liu Yanglet acknowledge the support from Columbia's SIRS and STAR Program, The Tang Family Fund for Research Innovations in FinTech, Engineering, and Business Operations. Shengyuan Colin Lin and Xiao-Yang Liu Yanglet acknowledge the support from NSF IUCRC CRAFT center research grant (CRAFT Grant 22017) for this research. The opinions expressed in this publication do not necessarily represent the views of NSF IUCRC CRAFT. 

\bibliography{custom}

\appendix
\onecolumn

\section{Related Work}
\label{sec:sec_related_work}

\subsection{Trend of Benchmark Development}

With the rapid development of large language models, many traditional benchmark datasets have become insufficiently challenging to effectively distinguish between models of varying capacities \cite{ivanov2025resurrecting}. Even relatively small-scale models now achieve comparable performance to much larger ones on these benchmarks, despite notable disparities in their real-world capabilities \cite{hsieh2023distilling}. 
The FinBen benchmark \cite{xie2024finben} also reveal that certain datasets are unable to clearly distinguish differences in model capabilities by conducting thorough experiments and evaluates LLMs across diverse financial tasks.

This performance saturation has motivated increasing efforts to design more difficult and nuanced evaluation tasks across various domains \cite{he2024olympiadbench, wang2024mmlu, white2024livebench}.
In the general domain, the BIG-Bench dataset \cite{suzgun2022challenging} compiles challenging tasks to stress-test LLMs. Subsequent work examines whether prompting techniques can enhance performance in such tasks.
In the medical domain, Chen et al. \cite{chen2025benchmarking} developed QA tasks requiring clinical reasoning and explanation, providing a more rigorous evaluation of LLMs' capabilities in healthcare.
In the financial domain, Zhao et al. \cite{zhao2023docmath} introduced datasets of varying difficulty. For simpler tasks, small models perform comparably to larger ones. However, for complex benchmarks like Complong, involving multi-table reasoning and long document processing, smaller models struggle while larger models maintain a lead. This highlights the need for more discriminative benchmarks to expose nuanced differences in model capabilities.

Building on these observations and the proliferation of financial datasets, some recent work has proposed criteria to filter and prioritize representative datasets for benchmarking \cite{reuel2024betterbench, cao2025should}. These efforts assess benchmarks based on implementation, design, documentation, maintenance, and construction to determine their effectiveness.
However, these approaches do not systematically evaluate the impact of dataset difficulty or establish objective criteria for filtering datasets based on complexity.  

To address this gap, we propose a strategy that dynamically filters datasets and prioritizes those that are both challenging and effective in evaluating the performance of different LLMs. This approach enables more targeted benchmarking by focusing on datasets that can better reveal meaningful differences between models.

\subsection{Multilingual Textual Financial Benchmarks}

Despite the growing number of financial benchmarks developed to assess the capabilities of large language models in domain-specific tasks, most existing benchmarks remain monolingual, predominantly focusing on English or Chinese. Notable examples include PIXIU \cite{xie2023pixiu}, CFBenchmark \cite{lei2023cfbenchmark}, FinanceBench \cite{islam2023financebench}, and BizBench \cite{koncel2023bizbench}, each offering tasks such as question answering, text classification, and numerical reasoning, all rooted in financial documents or news within a single language setting.
Some efforts have extended financial benchmarking to additional languages. For instance, \cite{peng2025plutus} and \cite{hirano2024construction} introduced benchmarks that encompass Greek and Japanese, broadening the linguistic focus beyond English and Chinese.
Although works like \cite{xie2024finben} and \cite{zhang2024dolares} expand beyond English by including tasks in both English and Spanish, they treat the two languages as separate tasks rather than integrating them into a unified multilingual evaluation. This approach limits the ability to assess cross-lingual generalization within financial contexts.

Some recent studies have attempted to expand financial benchmarks by incorporating bilingual tasks. For example, DolFin \cite{nakhle2025dolfin} includes English-Chinese financial QA and reasoning tasks. However, its focus remains limited to translation-based tasks, which do not fully reflect real-world multilingual applications.

Currently, there is a lack of comprehensive financial benchmarks that simultaneously encompass monolingual, bilingual, and multilingual settings. Most existing resources either focus solely on monolingual tasks or limited bilingual tasks, without integrating them into a unified multilingual framework. This gap highlights the need for more inclusive and versatile evaluation resources that can assess model performance across diverse linguistic settings.
Our work addresses this limitation by proposing a benchmark that not only evaluates LLMs across a wider range of financial tasks, but also considers dataset value filtering and cross-lingual robustness, setting the stage for more comprehensive and discriminative benchmarking in the financial domain.

\subsection{Multimodal Financial Benchmarks}  

Most financial benchmarks focus solely on text, with limited exploration of visual and audio data. Recent works like \cite{qiu2024evaluation} address financial document parsing through OCR, while FinAudio \cite{cao2025finaudiobenchmarkaudiolarge} targets financial audio content such as earnings calls. However, a unified evaluation framework that integrates all three modalities remains absent.  
Our work introduces a benchmark that incorporates text, visual, and audio data, emphasizing two key directions: (1) dynamic dataset filtering to prioritize challenging content and (2) promoting multilingual and multimodal benchmarks to reflect real-world financial complexities.  

For visual modality, existing financial datasets such as CORD~\cite{park2019cord} and FUNSD~\cite{jaume2019funsd} focus on semi-structured documents but lack the complexity of financial statements, FinMME~\cite{luo-etal-2025-finmme} and MME-Finance~\cite{gan2024mmefinancemultimodalfinancebenchmark} focus on image understanding rather than the complex structure of document-based financial information. Applied systems like nvIngest~\cite{nvIngest2024} extract tabular data from PDFs but are not research benchmarks.  
Modern multimodal LLMs, such as LayoutLMv2~\cite{xu2021layoutlmv2}, effectively combine text, layout, and visual cues but focus mainly on entity extraction rather than complete document conversion. Our benchmark targets the end-to-end transformation of scanned documents into structured formats like HTML, aligning with practical financial analytics needs.  

AudioLLMs such as AudioGPT~\cite{huang2024audiogpt} and AudioPaLM~\cite{rubenstein2023audiopalm} excel in spoken content processing. Financial-specific benchmarks are scarce, with FinAudio~\cite{cao2025finaudiobenchmarkaudiolarge} being the first to focus on financial audio but lacking integration with text and visual data.  
Our benchmark unifies text, visual, and audio modalities, enabling comprehensive evaluation of multimodal financial reasoning with metrics that assess both accuracy and semantic consistency across modalities.

\newpage
\section{Task Categories}
\label{sec:sec_task_categories}

Inspired by FinBen \cite{xie2024finbenholisticfinancialbenchmark}, we organize candidate datasets under a unified taxonomy of seven core financial NLP tasks:
\begin{itemize}
    \item \textit{Information Extraction (IE)} focuses on converting unstructured financial text into structured outputs.
    \item \textit{Textual Analysis (TA)} assesses a model’s ability to interpret sentiment, topic, or tone in financial discourse.
    \item \textit{Question Answering (QA)} evaluates comprehension of financial content through question answering.
    \item \textit{Text Generation (TG)} focuses on producing coherent, informative, and factually accurate financial text.
    \item \textit{Risk Management (RM)} targets detection or analysis of risk-related signals.
    \item \textit{Forecasting (FO)} measures a model’s ability to predict market trends or investor behavior.
    \item \textit{Decision-Making (DM)} simulates complex financial decision processes.
\end{itemize}

\section{Evaluation Models}
\label{sec:sec_models}

\begin{table*}[htbp]
\centering
\scriptsize
\begin{threeparttable}
\begin{adjustbox}{width=\textwidth}
\begin{tabular}{lllllll}
\toprule
\textbf{Access} & \textbf{Modality} & \textbf{Language} & \textbf{Domain} & \textbf{Target Language} & \textbf{Model} & \textbf{MOF Class} \\
\midrule
\multirow{2}{*}{Closed} & \multirow{1}{*}{Multimodal} & \multirow{1}{*}{Multilingual} & General & English & GPT-4o \cite{hurst2024gpt} & Class III - Open Model - In progress (17\%) \\
\cmidrule(lr){2-7}
& \multirow{1}{*}{Text} & \multirow{1}{*}{Multilingual} & General & English & o3-mini \cite{openaio3mini} & Class III - Open Model - In progress (0\%) \\
\midrule
\multirow{20}{*}{Open} & \multirow{4}{*}{Multimodal} & \multirow{4}{*}{Multilingual} & General & English & meta-llama/Llama-4-Scout-17B-16E-Instruct \cite{llama4meta} & Class III - Open Model - In progress (0\%) \\
& & & General & English & google/gemma-3-4b-it \cite{gemmateam2025gemma3technicalreport} & Class III - Open Model - In progress (17\%) \\
& & & General & English & google/gemma-3-27b-it \cite{gemmateam2025gemma3technicalreport} & Class III - Open Model - In progress (17\%) \\
& & & General & English & Qwen/Qwen2.5-Omni-7B \cite{Qwen2.5-Omni} & Class III - Open Model - In progress (67\%) \\
\cmidrule(lr){2-7}
& \multirow{8}{*}{Text} & \multirow{7}{*}{Multilingual} & General & English & meta-llama/Llama-3.1-70B-Instruct \cite{dubey2024llama} & Class III - Open Model - In progress (0\%) \\
& & & General & English & Deepseek-V3 \cite{liu2024deepseek} & Class III - Open Model - In progress (17\%) \\
& & & General & English & Qwen/Qwen2.5-32B-Instruct \cite{qwen2.5} & Class III - Open Model - In progress (33\%) \\
& & & General & Japanese & cyberagent/DeepSeek-R1-Distill-Qwen-32B-Japanese \cite{cyberagent-deepseek-r1-distill-qwen-32b-japanese} & Class III - Open Model - In progress (50\%)\tnote{1} \\
& & & Financial & Chinese & Duxiaoman-DI/Llama3.1-XuanYuan-FinX1-Preview \cite{xuanyuan2024} & Class III - Open Model - In progress (0\%)\tnote{1} \\
& & & Financial & Spanish & TheFinAI/FinMA-ES-Bilingual \cite{10.1145/3637528.3671554} & Class III - Open Model - In progress (50\%)\tnote{1} \\
& & & Financial & Greek & TheFinAI/plutus-8B-instruct \cite{peng2025plutus} & Class III - Open Model - In progress (17\%)\tnote{1} \\
\cmidrule(lr){3-7}
& & English & Financial & English & TheFinAI/finma-7b-full \cite{xie2023pixiu} & Class III - Open Model - In progress (67\%)\tnote{1} \\
\cmidrule(lr){2-7}
% & \multirow{3}{*}{Vision} & \multirow{2}{*}{Multilingual} & General & English & Qwen/Qwen-VL-Max \cite{Qwen-VL} & Class III - Open Model - In progress (0\%) \\
& \multirow{2}{*}{Vision} & Multilingual & General & English & Deepseek-VL-7B-Chat \cite{lu2024deepseekvl} & Class III - Open Model - In progress (17\%) \\
\cmidrule(lr){3-7}
& & English & General & English & llava-hf/llava-v1.6-vicuna-13b-hf \cite{liu2023improved} & Class III - Open Model - Qualified \& Class II - Open Tooling - In progress (80\%)\tnote{1} \\
\cmidrule(lr){2-7}
& \multirow{5}{*}{Audio} & \multirow{5}{*}{Multilingual} & General & English & Whisper-V3 \cite{pmlr-v202-radford23a} & Class III - Open Model - In progress (67\%) \\
& & & General & English & Qwen2-Audio-7B \cite{chu2024qwen2audiotechnicalreport} & Class III - Open Model - In progress (67\%) \\
& & & General & English & Qwen2-Audio-7B-Instruct \cite{chu2024qwen2audiotechnicalreport} & Class III - Open Model - In progress (67\%) \\
& & & General & English & SALMONN-7B \cite{tang2024salmonngenerichearingabilities} & Class III - Open Model - In progress (50\%)\tnote{1} \\
& & & General & English & SALMONN-13B \cite{tang2024salmonngenerichearingabilities} & Class III - Open Model - In progress (50\%)\tnote{1} \\
\bottomrule
\end{tabular}
\end{adjustbox}
% \vspace{0.5em}
\begin{tablenotes}
\tiny
\item[1] The model is fine-tuned from other organizations' models, and its MOF class is evaluated only on the fine-tuned portion.
\end{tablenotes}
% $^*$ means the model is fine-tuned on other organizations' models, and its MOF class is evaluated only for the fine-tuned part.
\end{threeparttable}
\caption{Overview of evaluated models with access type, modality, language scope, and MOF class.}
\label{tab:tab_models}
\end{table*}

\newpage
\section{Structured Difficulty-Aware Benchmarking of English Datasets}
\label{sec:sec_benchmarking}

\begin{center}
\begin{table}[htbp]
\centering
\scriptsize
\setlength{\textfloatsep}{5pt}
\begin{adjustbox}{max width=\textwidth}
\begin{tabular*}{\textwidth}{@{\extracolsep{\fill}}lllcccc}
\toprule
\textbf{Difficulty} & \textbf{Task} & \textbf{Dataset} & \textbf{GPT-4o} & \textbf{LLaMA3.1-70B-Instruct} & \textbf{Mean} & \textbf{Variance} \\
\midrule
\multirow{14}{*}{\textcolor{softgreen}{Easy}} & IE & EMIR-NER & 93.00 & 92.00 & 92.50 & 1.00 \\
& IE & SC & 88.00 & 87.00 & 87.50 & 1.00 \\
& TA & MA & 80.00 & 84.00 & 82.00 & 4.00 \\
& TA & FPB & 83.00 & 79.00 & 81.00 & 4.00 \\
& TA & FiQA-SA & 77.00 & 74.00 & 75.50 & 3.00 \\
& TA & Headlines & 78.00 & 60.00 & 69.00 & 18.00 \\
& TA & MultiFin & 66.00 & 69.00 & 67.50 & 3.00 \\
& DM & GOOG & 67.76 & 64.79 & 66.27 & 2.97 \\
& TA & FinArg-ACC & 67.00 & 65.00 & 66.00 & 2.00 \\
& TA & FOMC & 67.00 & 64.00 & 65.50 & 3.00 \\
& QA & XBRL-Math & 68.00 & 62.00 & 65.00 & 6.00 \\
% & DM & GM & 41.23 & 83.04 & 62.13 & 41.81 \\
& DM & AMZN & 62.22 & 59.01 & 60.62 & 3.22 \\
& DM & NFLX & 54.32 & 65.86 & 60.09 & 11.54 \\
\midrule
\multirow{29}{*}{\textcolor{softorange}{Medium}} & RM & polish & 58.50 & 57.00 & 57.75 & 1.5 \\
& TA & FinArg-ARC & 60.00 & 55.00 & 57.50 & 5.00 \\
& RM & LendingClub & 57.00 & 56.50 & 56.75 & 0.50 \\
& DM & MSFT & 41.32 & 71.05 & 56.18 & 29.73 \\
& RM & CCFraud & 52.50 & 56.50 & 54.50 & 4.00 \\
& RM & German & 50.00 & 55.50 & 52.75 & 5.50 \\
& RM & travelinsurance & 54.50 & 50.00 & 52.25 & 4.50 \\
& QA & LegalQA & 57.00 & 47.00 & 52.00 & 10.00 \\
& RM & taiwan & 50.00 & 53.00 & 51.50 & 3.00 \\
& TA & TSA & 61.00 & 42.00 & 51.50 & 19.00 \\
& RM & CCF & 52.50 & 50.00 & 51.25 & 2.50 \\
& FO & ACL18 & 50.00 & 51.50 & 50.75 & 1.50 \\
& DM & COIN & 45.18 & 56.25 & 50.71 & 11.07 \\
& RM & ProtoSeguro & 50.00 & 50.50 & 50.25 & 0.50 \\
& FO & BigData22 & 48.50 & 50.00 & 49.25 & 1.50 \\
& FO & CIKM18 & 47.50 & 51.00 & 49.25 & 3.50 \\
& IE & FINER-ORD & 78.00 & 18.00 & 48.00 & 60.00 \\
& RM & Australian & 50.00 & 42.50 & 46.25 & 7.50 \\
& DM & TSLA & 49.47 & 41.92 & 45.69 & 7.54 \\
& QA & MOFQA & 39.50 & 41.00 & 40.25 & 1.50 \\
& TA & MLESG & 36.00 & 44.00 & 40.00 & 8.00 \\
& QA & CDMQA & 37.00 & 39.00 & 38.00 & 2.00 \\
& DM & AAPL & 40.51 & 30.03 & 35.27 & 10.49 \\
& DM & NIO & 24.96 & 38.18 & 31.57 & 13.22 \\
& TA & FinABB & 31.00 & 25.00 & 28.00 & 6.00 \\
& TG & XBRL-EXP & 26.00 & 26.00 & 26.00 & 0.00 \\
& DM & DIS & 27.11 & 24.04 & 25.57 & 3.07 \\
& QA & TATQA & 0.00 & 44.00 & 22.00 & 44.00 \\
& TG & EDTSUM & 25.00 & 18.00 & 21.50 & 7.00 \\
\midrule
\multirow{9}{*}{\textcolor{red}{Hard}} & QA & XBRL-Dom & 6.00 & 8.00 & 7.00 & 2.00 \\
& IE & NER & 6.00 & 5.00 & 5.50 & 1.00 \\
& QA & FinQA & 5.00 & 0.00 & 2.50 & 5.00 \\
& IE & FinRED & 3.00 & 0.00 & 1.50 & 3.00 \\
& IE & CD & 0.00 & 0.00 & 0.00 & 0.00 \\
& IE & FNXL & 0.00 & 0.00 & 0.00 & 0.00 \\
& IE & FSRL & 0.00 & 0.00 & 0.00 & 0.00 \\
& TG & ECTSUM & 0.00 & 0.00 & 0.00 & 0.00 \\
& QA & ConvFinQA & 0.00 & 0.00 & 0.00 & 0.00 \\
\bottomrule
\end{tabular*}
\end{adjustbox}
\caption{Structured difficulty-aware benchmarking of English datasets. Datasets are ranked by their mean standardized performance, with selected datasets highlighted in bold.}
\label{tab:finben-difficulty}
\end{table}
\end{center}

\begin{center}
\includegraphics[width=\textwidth]{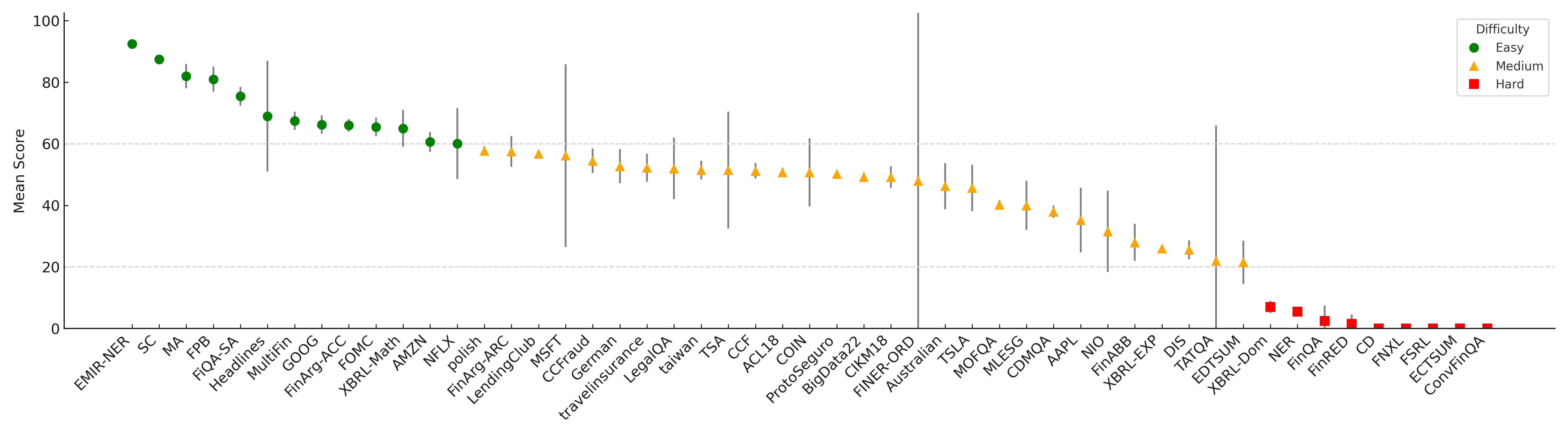}
\captionof{figure}{Structured difficulty-aware benchmarking of English datasets.}
\label{fig:finben-difficulty}
\end{center}

\newpage
\section{Textual Monolingual Benchmarks}
\label{sec:sec_textual_monolingual_benchmarks}

\subsection{English}
To comprehensively assess the understanding capabilities of LLMs in English regulatory and financial contexts, we collected 13 datasets for 7 tasks including information extraction (IE), question answering (QA), text analysis (TA), text generation (TG), risk management (RM), forecasting (FO), and decision-making (DM). All task data are sourced from diverse real-world scenarios, including the U.S. Securities and Exchange Commission (SEC), regulatory reporting, financial reporting, financial news, earnings call transcripts, and microblog.

For the IE task, it focuses on extracting structured financial information from unstructured text. We evaluate this capability using three datasets: SC~\cite{mariko-etal-2020-financial}, FinRED~\cite{sharma2022finred}, and FINER-ORD~\cite{shah2023finer}. The \textbf{SC}~\cite{mariko-etal-2020-financial} dataset assesses the detection of causal relationships in financial news and SEC filings, using F1~\cite{sokolova2009systematic} as the evaluation metric. The \textbf{FinRED}~\cite{sharma2022finred} dataset targets relation extraction, identifying relationships such as ``product/material produced'' and ``manufacturer'' in financial news and earnings call transcripts, also evaluated using F1~\cite{sokolova2009systematic}. The \textbf{FINER-ORD}~\cite{shah2023finer} dataset involves named entity recognition, extracting entities like organizations, locations, and persons from financial agreements and SEC filings, with performance measured by entity F1~\cite{derczynski-2016-complementarity}. Overall, the IE task provides a foundation for evaluating the model’s understanding of financial knowledge.

For the TA task, it encompasses tasks such as sentiment classification, topic detection, and event identification from financial texts. In this work, we focus on two representative financial datasets: Headlines~\cite{sinha2021impact} and TSA~\cite{cortis2017semeval}. The \textbf{Headlines}~\cite{sinha2021impact} dataset targets the extraction of actionable signals—such as price movements—from financial news headlines, with performance measured by the average F1 score (avg F1). The \textbf{TSA}~\cite{cortis2017semeval} dataset centers on sentiment analysis, where the goal is to classify textual segments into positive, negative, or neutral sentiment categories, evaluated using the accuracy~\cite{makridakis1993accuracy} metric. This task is designed to assess the capability of LLMs in performing fine-grained understanding and reasoning over domain-specific financial language.

For the QA task, it involves answering specific regulatory and financial questions by leveraging the inherent knowledge of LLMs. Both \textbf{FinQA}~\cite{chen2021finqa} and \textbf{TATQA}~\cite{zhu2021tat} assess models' numerical reasoning abilities using financial reports, tables, and contexts, whereas \textbf{XBRL-Math}~\cite{chen2022convfinqa} focus on equation inference. All are evaluated by accuracy~\cite{makridakis1993accuracy}. 

For the TG task, it involves summarizing or explaining specific content within regulatory or financial contexts, evaluating models' ability to produce coherent and informative text. \textbf{EDTSUM}~\cite{xie2023pixiu} targets the summarization of financial news articles into concise and informative summaries. In addition, we utilize the \textbf{ECTSUM}~\cite{mukherjee2022ectsum} dataset, which focuses on summarizing earnings call transcripts. Model performance on these summarization tasks is evaluated using ROUGE-1~\cite{lin-2004-rouge}.

For the RM task, it focuses on identifying, extracting, and analyzing risk-related information, interpreting numerical data, and evaluating a model's ability to understand complex relationships. Specifically, we use the CCF~\cite{feng2024empowering} datasets to assess LLMs' ability in our paper. The \textbf{CCF}~\cite{feng2024empowering} dataset is suitable for detecting fraud, which identifies whether transactions are ``fraudulent'' or ``non-fraudulent''. This dataset is evaluated by Matthews Correlation Coefficient (MCC)~\cite{matthews1975comparison}.

For the FO task, it evaluates a model’s ability to predict future market trends and investor behavior based on patterns in financial data. We focus on stock movement prediction, aiming to forecast whether a stock’s price will rise or fall using historical price data and social media signals from the \textbf{BigData22}~\cite{soun2022accurate} dataset. Model performance is assessed using MCC~\cite{matthews1975comparison}.

For the DM task, it evaluates an LLM’s ability to synthesize diverse financial information to formulate and execute trading strategies—an inherently challenging task even for human experts. We refer to \textbf{MSFT}~\cite{yu2024finconsynthesizedllmmultiagent} dataset, which consist of data over one year, simulating real-world trading scenarios using historical prices, news articles, and sentiment analysis. Model performance is assessed using Sharpe Ratio (SR)~\cite{sharpe1998sharpe}, providing a comprehensive evaluation of profitability, risk management, and strategic decision-making.

\subsection{Chinese}

To address domain-specific challenges in Chinese financial and regulatory contexts, we include 4 datasets from 3 core task types including IE, TA, and QA. These tasks require nuanced understanding of financial contexts, audit-related concepts, legal references, and inter-entity relationships, and serve to assess models' capabilities in financial reasoning, governance interpretation, and terminology grounding:

In IE task, \textbf{RRE} \cite{jiajia-etal-2024-auditwen} is a relation extraction task that identifies the semantic relation between two audit entities in a sentence. Relations span 8 predefined categories including \textit{audit issue, audit item, audit basis, audit method, audit institution, audit outcome, audited entity, and related sector or domain}. This task tests a model’s ability to infer structured relationships in financial text. % "审计问题", "审计事项", "审计依据", "审计方法", "审计机构", "审计成果", "被审计单位", and "涉及的行业或领域".

For TA task, \textbf{AIE} \cite{jiajia-etal-2024-auditwen} involves categorizing audit-target entities into 7 regulatory domains (\textit{fiscal audit, infrastructure audit, customs audit, financial audit, social security audit, tax audit, and environmental audit}). This task evaluates regulatory disambiguation within Chinese financial texts. % "财政审计", "公共工程审计", "海关审计", "金融审计", "社会保障审计", "税收审计", and "资源环保审计"
\textbf{LNE} \cite{jiajia-etal-2024-auditwen} is a legal grounding task where models must align a given audit scenario with the most relevant legal or regulatory category (\textit{financial regulations, fiscal law, personal income tax, financial supervision, labor and employment, listed companies, social security, industry regulation, value-added tax, tax administration, asset evaluation law, resource tax, integrated governance, and comprehensive tax policy}). It requires both regulatory reasoning and knowledge of Chinese legal terminology. % "财经法规", "财政法规", "个人所得税", "金融综合", "劳动就业", "上市公司", "社会保障", "行业管理", "增值税", "征收管理", "资产评估法规", "资源税", "综合管理", and "综合税收政策"

In QA task, \textbf{FinanceIQ} \cite{10.1145/3583780.3615285} is a multi-choice classification task where models need to answer Chinese-language financial domain-specific questions. The task evaluates an LLM’s understanding of domain-specific vocabulary, foundational financial knowledge, and contextual reasoning in Chinese.

All datasets are evaluated using classification accuracy~\cite{makridakis1993accuracy}.

\subsection{Japanese}

To capture the unique challenges posed by Japanese in financial contexts, we only find one public TA dataset from a Japanese Financial Benchmark \cite{hirano2024construction} specifically designed to evaluate Japanese financial language understanding.
 
For TA task, we adopt the \textbf{chabsa} dataset (chABSA: Aspect-Based Sentiment Analysis dataset in Japanese) \cite{kubo2018chabsa}, a binary classification task evaluates the sentiment polarity (positive or negative) of financial terms within securities reports. It is derived from publicly disclosed annual securities filings in Japan. The dataset includes annotated sentiment terms and is scored using the macro F1~\cite{hinojosa2024performance} metric, excluding neutral terms for evaluation consistency.

\subsection{Spanish}

To comprehensively evaluate the understanding capabilities of LLMs in Spanish regulatory and financial contexts, we included four domain-specific datasets from FLARE-ES \cite{10.1145/3637528.3671554} across three tasks: TA, QA, and TG.

For TA task, text classification and sentiment analysis were included. In text classification task, the \textbf{MultiFin}~\cite{jorgensen2023multifin} dataset evaluates the model’s ability to categorize Spanish financial texts. This dataset, which centers on Spanish headlines, contains 2,066 articles spanning six key financial categories: \textit{Business \& Management, Finance, Government \& Controls, Industry, Tax \& Accounting, and Technology}. The task challenges the model to accurately assign each headline to its respective sector, thereby testing its linguistic adaptability and domain-specific knowledge in Spanish.
In sentiment analysis task, the \textbf{TSA}~\cite{pan2023evaluation} dataset includes 3,892 entries from financial news and tweets, annotated to classify sentiments as \textit{positive, negative, or neutral}—capturing a nuanced spectrum of market emotions in Spanish. 

For QA task, the \textbf{EFPA} \footnote{\url{https://efpa-eu.org/}} dataset, derived from questions used in financial examinations by official examiner associations, present unique challenges.  
The EFPA dataset broadens this challenge with 228 questions featuring four answer choices each, covering a broader spectrum of financial topics including economic principles, fundamental financial concepts, and intricate calculations associated with financial products. This dataset is evaluated using accuracy~\cite{makridakis1993accuracy}.

For TG task, the \textbf{FNS-2023}~\cite{10386228} dataset, consisting of 232 Spanish annual reports from diverse financial companies. It aims to condense voluminous Spanish financial documents into succinct, informative abstracts, enhancing the accessibility and usability of financial information. Rich in detailed financial data and narratives, these reports pose a distinct challenge: distilling the essence of each document into a summary that captures the key information while preserving the factual integrity and coherence of the original text. This dataset is evaluated using ROUGE-1~\cite{lin-2004-rouge}. 

\subsection{Greek}

We adopt four core financial NLP datasets from Plutus-ben \cite{peng2025plutus}, the first Greek financial benchmark. These datasets span from IE, TA, QA, and TG tasks. 

For IE task, only GRFinNUM \cite{peng2025plutus} is included. \textbf{GRFinNUM} \cite{peng2025plutus} targets fine-grained classification of numerals into \textit{MONETARY, PERCENTAGE, TEMPORAL, QUANTITY, and OTHERS} categories, which captures the nuances of numerical semantics in long-form Greek financial texts. 
This dataset is evaluated using entity F1~\cite{derczynski-2016-complementarity}. 

For TA task, \textbf{GRMultiFin}~\cite{jorgensen-etal-2023-multifin} focuses on classifying concise Greek financial headlines into one of six predefined thematic categories: \textit{Business \& Management, Finance, Government \& Controls, Tax \& Accounting, Technology, and Industry}, emphasizing the need for lexical disambiguation and contextual inference due to the brevity and ambiguity typical of headlines. This dataset is evaluated using accuracy~\cite{makridakis1993accuracy}. 

For QA task, \textbf{GRFinQA} \cite{peng2025plutus} evaluates models' performance on inferring the correct answer using provided financial text under a multiple-choice format, assessing a model’s ability to comprehend and reason over finance-related questions drawn from Greek academic sources. This dataset is evaluated using accuracy~\cite{makridakis1993accuracy}. 

For TG task, \textbf{GRFNS-2023}~\cite{10386228} tests models' ability of generating fluent, coherent, and factually accurate summaries from complex narrative sections of Greek financial reports, testing a model’s capacity for paraphrasing and information compression. This dataset is evaluated using ROUGE-1~\cite{lin-2004-rouge}. 

\newpage
\section{Audio Benchmark}
\label{sec:sec_audio_benchmark}

The Automatic Speech Recognition (ASR) task evaluates AudioLLMs' accuracy in transcribing financial audio clips, directly impacting applications such as financial voice assistants. In this task, each input is represented as $(A, Q, R)$, where $A$ is an audio clip, $Q$ is the prompt instruction, and $R$ is the reference transcript. The transcribed text $T$ is generated as $T = \textit{AudioLLM}(A, Q)$. The Word Error Rate (WER)~\cite{park2008empirical} is computed as $\text{WER} = \frac{S + D + I}{N}$, where $S$, $D$, and $I$ represent the number of substitutions, deletions, and insertions, respectively, and $N$ is the total number of words in the reference ($N = S + D + C$, with $C$ indicating correct words). A lower WER indicates better ASR performance.

\begin{table*}[h]
\resizebox{0.98\linewidth}{!}{
\begin{tabular}{l|c|r|r|c|c}
\toprule
Dataset Name & Type        & \#Samples & \# Hours & Task                               & Metrics              \\ \midrule
MDRM-test~\cite{qin2019you}         & Short Clips & 22,208     & 87       & short financial clip ASR           & WER                  \\
% SPGISpeech-test   & Short Clips & 39,341     & 130      & short financial clip ASR           & WER                  \\
% Earning-21   & Long Audio  & 44        & 39       & long financial audio ASR           & WER                  \\
% Earning-22   & Long Audio  & 125       & 120      & long financial audio ASR           & WER                  \\
FinAudioSum~\cite{cao2025finaudiobenchmarkaudiolarge}  & Long Audio  & 64        & 55       & long financial audio Summarization & ROUGE-L \\ \bottomrule
\end{tabular}}
\caption{Statistics of the datasets in the FinAudio benchmark.}
\vspace{-0.1em}
\label{tab:audio}
\end{table*}

The evaluation dataset in FinAudio \cite{cao2025finaudiobenchmarkaudiolarge} is developed from two primary data sources: 1) existing open-source financial audio data originally created for non-LLM evaluation purposes. 2) novel datasets introduced in this work. Table~\ref{tab:audio} summarizes the key statistics.

\begin{itemize}
    \item Short financial audio clip dataset: 
    % Two datasets provide financial audio clips: 
    the MDRM~\cite{qin2019you} dataset derived from earnings conference call recordings. 
    % and SPGISpeech~\cite{o2021spgispeech}, 
 The original audio data was segmented at the sentence level and aligned with corresponding transcripts. This dataset is initially split into training and test subsets. For our paper, we utilize only the test sets for evaluation, which includes 22,208 audio clips totaling 87 hours. %while the SPGISpeech-test set contains 39,341 clips totaling 130 hours.
    
    % \item Long financial audio recording datasets: Two datasets provide long financial audio recordings: Earnings-21~\cite{del2021earnings}, containing 44 earnings calls (approximately 39 hours), and Earnings-22~\cite{del2022earnings}, with 125 earnings calls (120 hours). Each recording includes a corresponding manual transcript. We use both datasets entirely for evaluation.

    \item FinAudioSum: \textit{FinAudioSum} \cite{cao2025finaudiobenchmarkaudiolarge} was created based on the ECTSum dataset~\cite{mukherjee2022ectsum}, originally designed for earnings call summarization using textual data. ECTSum comprises 2,425 earnings transcripts paired with expert-generated, telegram-style summaries. We obtain corresponding audio recordings for the ECTSum test set from earningcast\footnote{\url{https://earningscast.com/}}. Overlapping recordings with Earnings-21 and Earnings-22 (spanning 2019–2022) are removed. The final \textit{FinAudioSum} \cite{cao2025finaudiobenchmarkaudiolarge} dataset includes 64 recordings totaling 55 hours.

\end{itemize}

\section{OCR Statistics}
\label{sec:sec_ocr_statistics}

\begin{table}[htbp]
\centering
\small
\setlength{\tabcolsep}{6pt}
\renewcommand{\arraystretch}{1.05}
\begin{threeparttable}
\begin{tabular}{lccc}
\toprule
\textbf{Dataset} & \textbf{\# PDF Files} & \textbf{Storage (GB)} & \textbf{Size} \\
\midrule
EnglishOCR  & 1,039 & 1.70 & 7,961 \\
JapaneseOCR & 331   & 1.57 & 17,586 \\
SpanishOCR  & 2,302 & 0.76 & 12,819 \\
GreekOCR    & 100   & 0.16 & 6,533 \\
\bottomrule
\end{tabular}
\caption{Statistics of OCR datasets.}
\label{tab:tab_ocr}
\end{threeparttable}
\end{table}

\newpage
\section{PolyFiQA-Easy and PolyFiQA-Expert}

\subsection{Data Construction}
\label{sec:data_construction}

\begin{figure}[h]
  \includegraphics[width=\columnwidth]{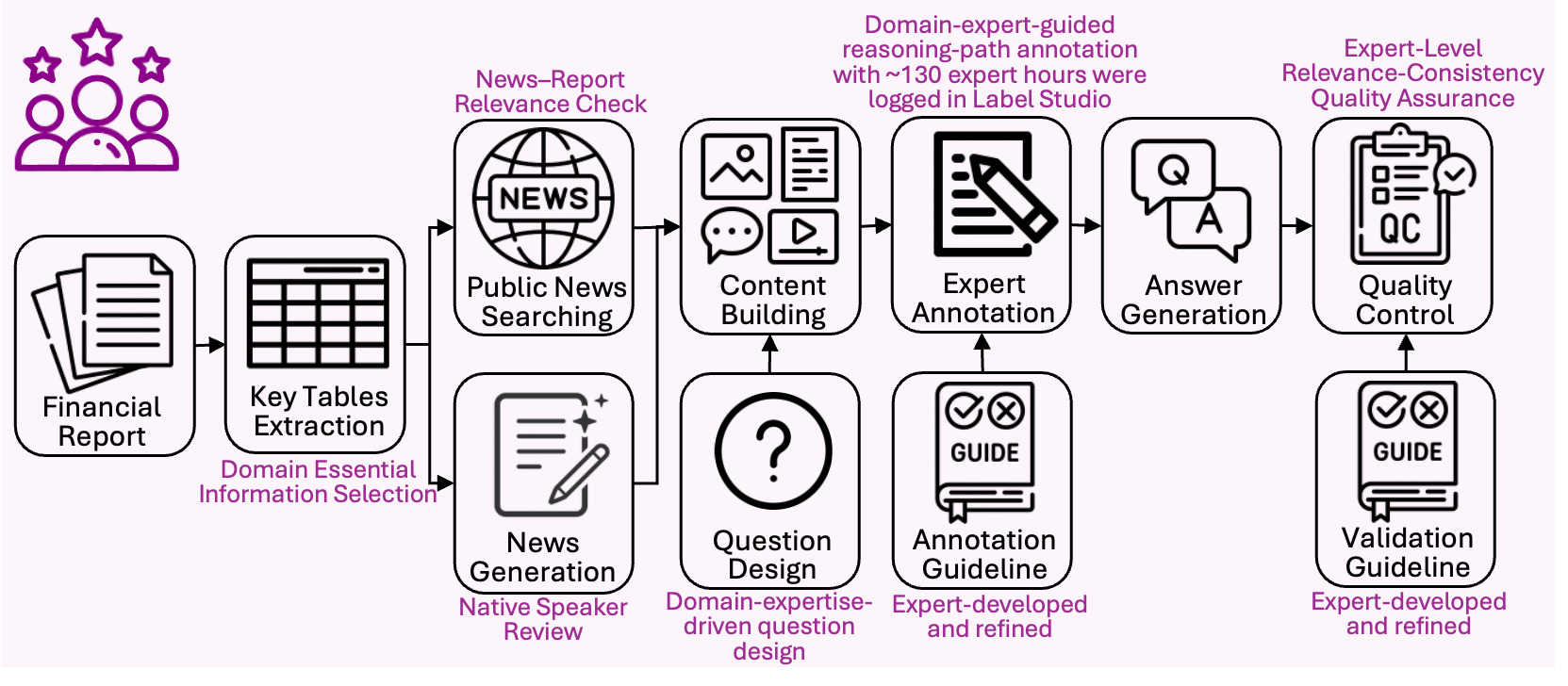}
  \caption{Expert-in-the-loop data construction of \textit{PolyFiQA-Easy} and \textit{PolyFiQA-Expert}.}
  \label{fig:fig_polyfiqa}
\end{figure}

\paragraph{Expert-in-the-Loop Data Construction.} 

To ensure benchmark fidelity and domain rigor, we adopt an expert-in-the-loop pipeline (Figure \ref{fig:fig_polyfiqa}). Three financial professionals (Appendix~\ref{sec:sec_annotator_demography}) with expertise in economics, business, and accounting oversaw all phases of construction, including news selection, question authoring, guideline development, annotation, and quality control.
News articles were meticulously screened for strong alignment with financial reports.
Questions were crafted to anchor in real analytical tasks and span two difficulty tiers: easier questions in \textit{PolyFiQA-Easy} and more complex ones in \textit{PolyFiQA-Expert} (Appendix~\ref{sec:sec_question_design}). This structure supports fine-grained model assessment across reasoning levels (Appendix~\ref{sec:sec_reasoning_trajectory}).
Rigorous tier-specific annotation guidelines (Appendix~\ref{sec:sec_annotator_guideline}) were refined through iterative pilot annotation rounds including tier-specific annotation rules and formatting protocols to promote inter-annotator consistency.
In total, approximately 130 expert hours were logged in Label Studio (Appendix~\ref{sec:sec_annotation_process}), establishing a high quality, reproducible, auditable, and streamlined workflow aligned with best practices in benchmark creation.
The raw datasets were further converted into structured instruction datasets with task-specific prompts thoughtfully crafted by financial professionals (Appendix~\ref{sec:sec_instruction}).

\paragraph{Quality Validation.}

Evaluating the quality of free-text generation datasets remains a persistent challenge in both QA and TG. To ensure annotation reliability, we adopt a structured scoring framework inspired by prior summarization benchmarks \cite{takeshita-etal-2024-aclsum, gliwa-etal-2019-samsum}, evaluating each instance along two key dimensions:
\textbf{Relevance} (scored 1–4) captures whether the response includes the key information required to answer the question.
\textbf{Consistency} (scored 1–3) measures factual accuracy, especially numerical values.
Each question is initially annotated by one expert and independently scored by two additional reviewers using detailed, pilot-refined validation guidelines (Appendix~\ref{sec:sec_validation_guideline}). Only responses with cumulative scores above 5 are retained in the final dataset. 
To further validate scoring reliability, we report inter-annotator agreement as normalized difference percentage across dimensions.
\textit{PolyFiQA-Easy} and \textit{PolyFiQA-Expert} achieved average inter-annotator agreements of 89.38\% and 91.21\%, demonstrating the benchmark’s high quality and scoring consistency.

\newpage
\subsection{News Data Sources}
\label{sec:sec_news_data_source}

We collected news data from the following resources:

\begin{itemize}
    \item \textbf{English:} Articles were primarily obtained from prior work \cite{yu2024finconsynthesizedllmmultiagent}; additional samples were generated by the authors.
    \item \textbf{Chinese:} Articles were collected from public websites including \url{https://guba.eastmoney.com/}, \url{https://caifuhao.eastmoney.com}, \url{https://www.tobaccochina.com}, \url{https://www.jiemian.com}, and \url{https://xueqiu.com}.
    \item \textbf{Japanese:} Articles were primarily collected from \url{https://news.mynavi.jp/}, \url{https://www3.nhk.or.jp}, \url{https://jp.investing.com}, and \url{https://sustainablejapan.jp}; additional samples were generated by the authors.
    \item \textbf{Spanish:} Articles were primarily collected from \url{https://www.diariolibre.com}, \url{https://www.abc.es}, \url{https://es.marketscreener.com}, \url{https://www.infobae.com/}, and \url{https://es.investing.com}; additional samples were generated by the authors.
    \item \textbf{Greek:} Since no publicly available Greek news sources were identified, all articles were generated by the authors.
\end{itemize}

\subsection{News Generation}
\label{sec:sec_news_generation}

To support high-quality, multilingual financial news creation in our benchmark, we designed a semi-automated pipeline that integrates human curation with LLM assistance. The pipeline consists of three main stages:

\paragraph{Step 1: Extracting key financial highlights.}
For each target company and reference date, we collect relevant financial news and the corresponding SEC filings (10-Q or 10-K). From these sources, we distill 2–3 bullet points summarizing key developments, with emphasis on material indicators such as revenue performance, profit margins, and liquidity or cash flow metrics.

\paragraph{Step 2: AI-assisted drafting.}
We prompt GPT-4o to generate a concise financial news article based on the extracted bullet points. The prompt enforces a neutral and professional journalistic tone, while discouraging unsupported statements:

\begin{quote}
\texttt{You are a journalist, specializing in finance. I am giving you some basic points, based on which I want you to generate a financial news article. This article should not be very large. Use neutral tone and professional language. Do not generate fallacious statements. You can change the order of the bullet points.}
\end{quote}

\paragraph{Step 3: Human post-editing and quality assurance.}
The AI-generated draft is subsequently reviewed by human annotators to improve readability, ensure adherence to journalistic standards, and remove any hallucinated or unverifiable content. This step guarantees that the final article is copyright-compliant and fully authored by the benchmark creators.

\newpage
\subsection{Question Design}
\label{sec:sec_question_design}

The questions are categorized into two difficulty levels: \textbf{Easy} and \textbf{Expert}.

\paragraph{PolyFiQA-Easy:}
\begin{itemize}
    \item What trends can be observed in the company’s revenue amount over the past few years? Please quote the piece(s) of news aligned with the finding(s), if any.
    \item How does the company’s balance sheet reflect its financial health in terms of total amount of current assets and the ratio of total liability divided by total equity? Please provide only the final result(s); do not include any calculation steps. Please quote the piece(s) of news aligned with the finding(s), if any.
    \item Are there any significant changes or irregularities in the amount of operating, investing and financing cash flow? Please quote the piece(s) of news aligned with the finding(s), if any.
    \item What is the company's R\&D ratio (R\&D divided by revenue)? Please provide only the final result(s); do not include any calculation steps. Please quote the piece(s) of news aligned with the finding(s), if any.
\end{itemize}

\paragraph{PolyFiQA-Expert:}
\begin{itemize}
    \item Please list the top three focuses on revenue from news. Please quote all the relevant financial information from the financial statements supporting the finding(s), if any.
    \item How is the company allocating capital (e.g., investments, share repurchases, dividends) based on its performance and market outlook from the news? Please quote all the relevant financial information from the financial statements supporting the finding(s), if any.
    \item What is the company’s strategy on maintaining profit margins from the news? Please quote all the relevant financial information from the financial statements supporting the finding(s), if any.
    \item What are the company's capital expenditures and their strategic significance from the news? Please quote all the relevant financial information from the financial statements supporting the finding(s), if any.
\end{itemize}

\subsection{Reasoning Trajectory}
\label{sec:sec_reasoning_trajectory}

\paragraph{PolyFiQA-Easy:}
\begin{itemize}
    \item \textbf{Q:} What trends can be observed in the company’s revenue amount over the past few years? Please quote the piece(s) of news aligned with the finding(s), if any.
    
    \textbf{Steps:} 
    \begin{enumerate}
        \item Extract revenue amounts over the past few years.
    \end{enumerate}

    \item \textbf{Q:} How does the company’s balance sheet reflect its financial health in terms of total amount of current assets and the ratio of total liability divided by total equity? Please provide only the final result(s); do not include any calculation steps. Please quote the piece(s) of news aligned with the finding(s), if any.
    
    \textbf{Steps:} 
    \begin{enumerate}
        \item Extract current assets, total liability, and total equity amounts.
        \item Calculate the ratio (total liability divided by total equity).
    \end{enumerate}

    \item \textbf{Q:} Are there any significant changes or irregularities in the amount of operating, investing and financing cash flow? Please quote the piece(s) of news aligned with the finding(s), if any.
    
    \textbf{Steps:} 
    \begin{enumerate}
        \item Extract all three cash flow components (operating, investing and financing cash flow amounts).
    \end{enumerate}

    \item \textbf{Q:} What is the company's R\&D ratio (R\&D divided by revenue)? Please provide only the final result(s); do not include any calculation steps. Please quote the piece(s) of news aligned with the finding(s), if any.
    
    \textbf{Steps:} 
    \begin{enumerate}
        \item Extract R\&D and revenue amounts.  
        \item Calculate R\&D ratio (R\&D amount divided by revenue amount).
    \end{enumerate}
\end{itemize}

\paragraph{PolyFiQA-Expert:}
\begin{itemize}
    \item \textbf{Q:} Please list the top three focuses on revenue from news. Please quote all the relevant financial information from the financial statements supporting the finding(s), if any. 
    
    \textbf{Steps:} 
    \begin{enumerate}
        \item Extract revenue for all focuses.  
        \item Rank the revenue amounts of different focuses from high to low.  
        \item Select the top three.
    \end{enumerate}

    \item \textbf{Q:} How is the company allocating capital (e.g., investments, share repurchases, dividends) based on its performance and market outlook from the news? Please quote all the relevant financial information from the financial statements supporting the finding(s), if any.
    
    \textbf{Steps:} 
    \begin{enumerate}
        \item Extract capital allocation items (investments, dividends, etc). 
        \item Extract market outlook information from news.
        \item Analyze capital amounts and market outlook information together.
    \end{enumerate}

    \item \textbf{Q:} What is the company’s strategy on maintaining profit margins from the news? Please quote all the relevant financial information from the financial statements supporting the finding(s), if any.
    
    \textbf{Steps:} 
    \begin{enumerate}
        \item Extract net profit and revenue.  
        \item Calculate profit margin (net profit divided by revenue amount).  
        \item Extract strategy information.  
        \item Analyze profit margin ratio and strategy information.
    \end{enumerate}

    \item \textbf{Q:} What are the company's capital expenditures and their strategic significance from the news? Please quote all the relevant financial information from the financial statements supporting the finding(s), if any.
    
    \textbf{Steps:} 
    \begin{enumerate}
        \item Extract capital expenditures.  
        \item Extract strategic context.  
        \item Analyze capital expenditures amounts and strategy information together.
    \end{enumerate}
\end{itemize}

\newpage
\subsection{Annotation Guideline}
\label{sec:sec_annotator_guideline}
\subsubsection{Task Description}

The task is a question answering (QA) and text generation (TG) task based on both financial statements and financial news. 

\subsubsection{Annotation Rules}
\paragraph{General rules}
\begin{itemize}
    \item Limit your response to 100 words or fewer.
    \item Write \texttt{None} if the question cannot be answered or if no relevant evidence is found.
\end{itemize}

\paragraph{Specific rules}

\subparagraph{(1) PolyFiQA-Easy:}
\begin{itemize}
    \item \textbf{Answer format:}
    \begin{itemize}
        \item \textbf{Answer:} \{Answer using the financial statement.\}
        \item \textbf{News Evidence:} \{Verify your answer using quote(s) from financial news. Write \texttt{None} if no news evidence is available.\}
    \end{itemize}
\end{itemize}

\subparagraph{(2) PolyFiQA-Expert:}
\begin{itemize}
    \item \textbf{Answer format:}
    \begin{itemize}
        \item \textbf{Answer:} \{Answer by summarizing the financial news.\}
        \item \textbf{Financial Statements Evidence:} \{Verify your answer using quote(s) from the financial statements (include original amounts if relevant). Write \texttt{None} if no statement evidence is available.\}
    \end{itemize}
\end{itemize}

\subsection{Annotator Demography}
\label{sec:sec_annotator_demography}

The construction of our \textsc{\multifinben} relies on the deep domain expertise of a team of highly qualified annotators with strong backgrounds in finance, economics, and data science. Their interdisciplinary training and multilingual fluency ensure accurate and contextually grounded annotations across both financial statements and cross-lingual news articles. % This is especially critical for our dataset, which spans multiple filings across companies, each annotated with eight analytically demanding questions that require close reading, financial reasoning, and linguistic nuance.

One annotator, currently working as a senior analyst at a major U.S. financial institution, holds a master’s degree in Business Analytics from a leading Ivy League university and a bachelor’s degree in Statistics and Economics. Their background includes research on LLMs, financial data analysis, and economics, enabling precise annotation of complex financial information. Fluent in Chinese and experienced in multilingual reasoning, they bring a high level of rigor to aligning financial data with multilingual news content.

Another annotator earned their master’s degree in Financial Mathematics from a prominent U.S. institution. They are currently pursuing a second master’s degree in Computer Science with a focus on machine learning at a major public research university in the United States. With over seven years of professional experience in strategic finance and consulting within the FinTech industry, they contribute practical expertise in corporate finance, along with a strong foundation in computational methods. 

The team is further strengthened by an annotator currently serving as a Research Associate at a major international financial research institute in Tokyo. They hold a Ph.D. in Financial Economics from a leading Australian university, with research spanning mergers and acquisitions, machine learning, and firm behavior. Their background includes prior academic positions in both Australia and China, multiple peer-reviewed publications in finance journals, and presentations at major international conferences. They also have experience as an editorial assistant, ad hoc referee, and research assistant, contributing a strong academic foundation to the annotation process.

% The team is further strengthened by an annotator currently serving as a Research Associate at the Asian Development Bank Institute (ADBI) in Tokyo. They hold a Ph.D. in Financial Economics from a leading Australian university, where their research focused on mergers and acquisitions, machine learning, and firm behavior. They also earned an M.Phil. in Finance and a Master’s in Accounting and Finance from the same institution. Their experience includes academic positions at universities in both Australia and China, along with multiple publications and conference presentations at venues such as the Pacific-Basin Finance Journal and the 2023 Chicago FMA Conference. They have additionally contributed to the field as an editorial assistant, ad hoc referee, and research assistant, bringing a deep academic foundation to the annotation process.

Together, the team’s combined financial acumen, analytical precision, and multilingual fluency enable the construction of a high-quality, cross-lingual financial QA dataset. Their expertise ensures the benchmark is linguistically, technically, and financially robust, setting a strong foundation for advancing multilingual LLM research in financial domain.

\newpage
\subsection{Annotation Process}
\label{sec:sec_annotation_process}

\begin{figure}[h]
  \label{fig:fig_annotation}
  \centering
  \includegraphics[width=\linewidth]{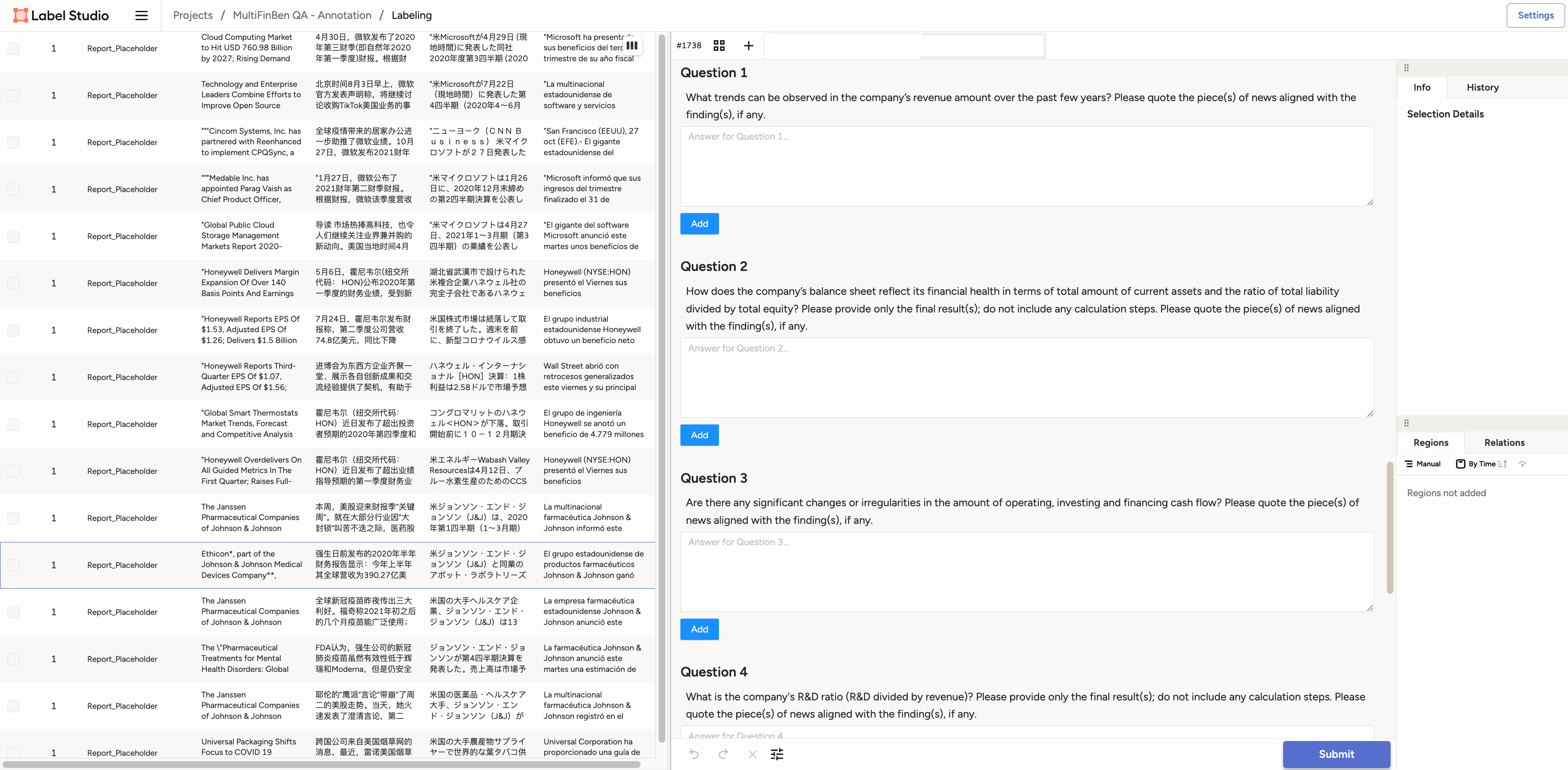}
  % \fbox{\rule[-.5cm]{0cm}{4cm} \rule[-.5cm]{4cm}{0cm}}
  \caption{The Label Studio interface of the PolyFiQA-Easy and PolyFiQA-Expert annotation process.}
\end{figure}

\subsection{Validation Guideline}
\label{sec:sec_validation_guideline}

To ensure consistent annotation quality in PolyFiQA-Easy and PolyFiQA-Expert, we define the following validation guidelines:

(1) \textbf{Relevance}: Assesses whether the response includes the key information needed to answer the question.
\begin{itemize}
    \item 4 – The response includes all key points or if no answer could be generated from the content.
    \item 3 – The response partially includes key points without irrelevant information.
    \item 2 – The response partially includes key points with minor irrelevant information.
    \item 1 – The response includes no relevant information.
\end{itemize}

(2) \textbf{Consistency (Factual accuracy)}: Evaluates the correctness of numerical values.
\begin{itemize}
    \item 3 – All numbers are correct or if no numbers are needed.
    \item 2 – Some numbers are incorrect.
    \item 1 – All numbers are incorrect.
\end{itemize}

\subsection{Annotation Agreement Metric}

To account for the annotation agreement in our consistency rating task, we report \textbf{Gwet's AC1} \cite{gwet2014handbook}, a chance-corrected agreement coefficient that is more robust to prevalence and marginal distribution issues than Cohen's Kappa. The AC1 statistic is defined as:
\begin{equation}
\mathrm{AC1} = \frac{P_o - P_e}{1 - P_e},
\end{equation}
where $P_o$ is the observed agreement, and $P_e$ is the chance agreement estimated by:
\begin{equation}
P_e = \sum_{i=1}^{k} p_i (1 - p_i),
\end{equation}
with $p_i$ being the proportion of annotations assigned to class $i$. This metric was selected due to the class imbalance present in our dataset.

\newpage
\subsection{Instruction Data Conversion}
\label{sec:sec_instruction}

\begin{tcolorbox}[colback=lightgray!10, colframe=black, title=Task Instruction for PolyFiQA-Easy]
\begin{verbatim}
You are tasked with answering the user's question using the provided 
context, which includes financial statements (Income Statements, Balance 
Sheets, and Cash Flow Statements) and financial news articles in 
multiple languages (English, Chinese, Japanese, Spanish, and Greek). 
Please provide a detailed and well-supported answer based on the 
information available. Limit your response to 100 words or fewer. If you 
cannot answer the question or if no relevant evidence is found, write 
“None”. 

Answer Format:
Answer: {Answer using the financial statement.}  
News Evidence: {Verify your answer using quote(s) from the financial 
news. Write “None” if no news evidence is available.}

Context:
Financial Statements:
{financial_statements}

English News:
{english_news_content}

Chinese News:
{chinese_news_content}

Japanese News:
{japanese_news_content}

Spanish News:
{spanish_news_content}

Greek News:
{greek_news_content}

Question:
{user_question}

Answer:
\end{verbatim}
\end{tcolorbox}

\begin{tcolorbox}[colback=lightgray!10, colframe=black, title=Task Instruction for PolyFiQA-Expert]
\begin{verbatim}
You are tasked with answering the user's question using the provided 
context, which includes financial statements (Income Statements, Balance 
Sheets, and Cash Flow Statements) and financial news articles in 
multiple languages (English, Chinese, Japanese, Spanish, and Greek). 
Please provide a detailed and well-supported answer based on the 
information available. Limit your response to 100 words or fewer. If you 
cannot answer the question or if no relevant evidence is found, write 
“None”. 

Answer Format:
Answer: {Answer by summarizing the financial news.}  
Financial Statements Evidence: {Verify your answer using quote(s) from 
the financial statements (include original amounts if relevant). Write 
“None” if no statement evidence is available.}

Context:
Financial Statements:
{financial_statements}

English News:
{english_news_content}

Chinese News:
{chinese_news_content}

Japanese News:
{japanese_news_content}

Spanish News:
{spanish_news_content}

Greek News:
{greek_news_content}

Question:
{user_question}

Answer:
\end{verbatim}
\end{tcolorbox}

\newpage
\section{Error Analysis}
\label{sec:sec_error_analysis}

\subsection{\textit{PolyFiQA-Easy} and \textit{PolyFiQA-Expert}}

Expert reviewers analyzed model outputs from GPT-4o and LLaMA3.1-70B-Instruct, identifying \textbf{information extraction failures} as the most common error source. Models often cited \textbf{incorrect financial fields} (e.g., using total assets instead of cash flow, or retained earnings instead of net profit margin), reflecting confusion over semantically similar numeric entries. In some cases, models returned ``none'' when they failed to locate relevant information. 

These errors indicate persistent challenges in \textbf{retrieving and interpreting structured financial data}, especially in multilingual and cross-document settings. The significance of this task lies in its real-world complexity: financial professionals routinely integrate structured filings and multilingual news to form investment insights. \textit{PolyFiQA-Easy} and \textit{PolyFiQA-Expert}, as the first benchmark to capture this \textbf{multilingual, cross-document reasoning challenge}, provides a high-fidelity testbed for evaluating and advancing LLMs toward realistic financial QA capabilities.

\subsection{\textit{EnglishOCR}, \textit{JapaneseOCR}, \textit{SpanishOCR}, and \textit{GreekOCR}}

Expert review of GPT-4o's outputs identified five major error types:
\begin{enumerate}
    \item \textbf{Omission of tables and charts}, especially in complex or multi-column layouts, leading to loss of key financial content;
    \item \textbf{Hallucinations}, where models generated plausible but ungrounded content;
    \item \textbf{Numeric misinterpretation}, including rounding errors, dropped digits, and misreading of formatting conventions;
    \item \textbf{Edge-based omissions}, where headers, footers, or marginal content was ignored, possibly due to incomplete OCR bounding or attention bias;
    \item \textbf{Skipping bracketed or vertical text}, which often contained essential qualifiers or context.
\end{enumerate}

These diverse and persistent error types underscore the real-world difficulty and importance of our OCR benchmark. Accurate, layout-aware extraction of financial content is essential for downstream tasks such as auditing, reporting, and compliance. With four novel OCR datasets (\textit{EnglishOCR}, \textit{JapaneseOCR}, \textit{SpanishOCR}, and \textit{GreekOCR}), our benchmark supports \textbf{fine-grained diagnosis} of model limitations and advances the frontier of multimodal financial NLP.

\newpage
\section{Human Evaluation of OCR Outputs}
\label{sec:sec_ocr_human_evaluation}

Automatic metrics such as ROUGE-1 may not fully capture \textbf{factual consistency} in long-form OCR outputs. To complement quantitative evaluation, we conducted a \textbf{manual factuality assessment} of GPT-4o's OCR results following a structured annotation guideline. Inter-annotator agreement (IAA) was computed to ensure reliability.

\subsection{Factuality Scoring Guideline}

\begin{itemize}
    \item \textbf{1 (Bad):} Multiple factual inaccuracies, such as incorrect company names, locations, or numerical data.
    \item \textbf{2 (Okay):} Generally accurate but contains omissions or minor inconsistencies in the extracted content.
    \item \textbf{3 (Good):} Mostly accurate with small formatting deviations, such as missed page numbers.
    \item \textbf{4 (Excellent):} Fully faithful to the source document, with all content and structure accurately preserved.
\end{itemize}

\subsection{Human Evaluation}

\begin{table}[htbp]
\centering
\small
\setlength{\tabcolsep}{8pt}
\renewcommand{\arraystretch}{1.1}
\begin{tabular}{lccc}
\toprule
\textbf{Dataset} & \textbf{Mean Factuality Score} & \textbf{Cohen’s $\kappa$} & \textbf{Krippendorff’s $\alpha$} \\
\midrule
EnglishOCR (n=256) & 2.224 & 0.752 & 0.841 \\
SpanishOCR (n=50) & 2.865 & 0.886 & 0.930 \\
GreekOCR (n=100) & 2.865 & 0.886 & 0.930 \\
\bottomrule
\end{tabular}
\caption{Human factuality evaluation of OCR outputs from GPT-4o.}
\label{tab:ocr_human_eval}
\end{table}

This evaluation highlights the limitations of automatic similarity-based metrics and demonstrates the need for \textbf{manual factual assessment} in layout-rich, domain-specific OCR tasks.

\newpage

\newpage
\section{Model Openness Framework (MOF)}
\label{sec:mof}

To systematically assess the openness and completeness of the models listed above, we adopt the Model Openness Framework (MOF) \cite{mof2024white}, a three-tier ranked classification system designed for machine learning models. The MOF defines 17 components spanning the model development lifecycle and categorizes models into three hierarchical classes: ``Class III - Open Model'', ``Class II - Open Tooling'', and ``Class I - Open Science'', with each level subsuming the requirements of the preceding one. Class III represents the minimal threshold of openness. Although many models are advertised as ``open'', few satisfy even the Class III criteria. Most release only partial artifacts, such as model architecture, weights, model cards, or technical reports, without permissive open-source licenses. Consequently, downstream use, modification, and redistribution are often legally constrained, posing potential legal and compliance risks. Among the models in our benchmark, most fall under Class III, with only llava-v1.6-vicuna-13b qualifying for Class II.

\section{Open Multilingual and Multimodal Financial LLM Leaderboard}
\label{sec:sec_leaderboard}

\begin{figure}[h]
  \centering
  \includegraphics[width=\linewidth]{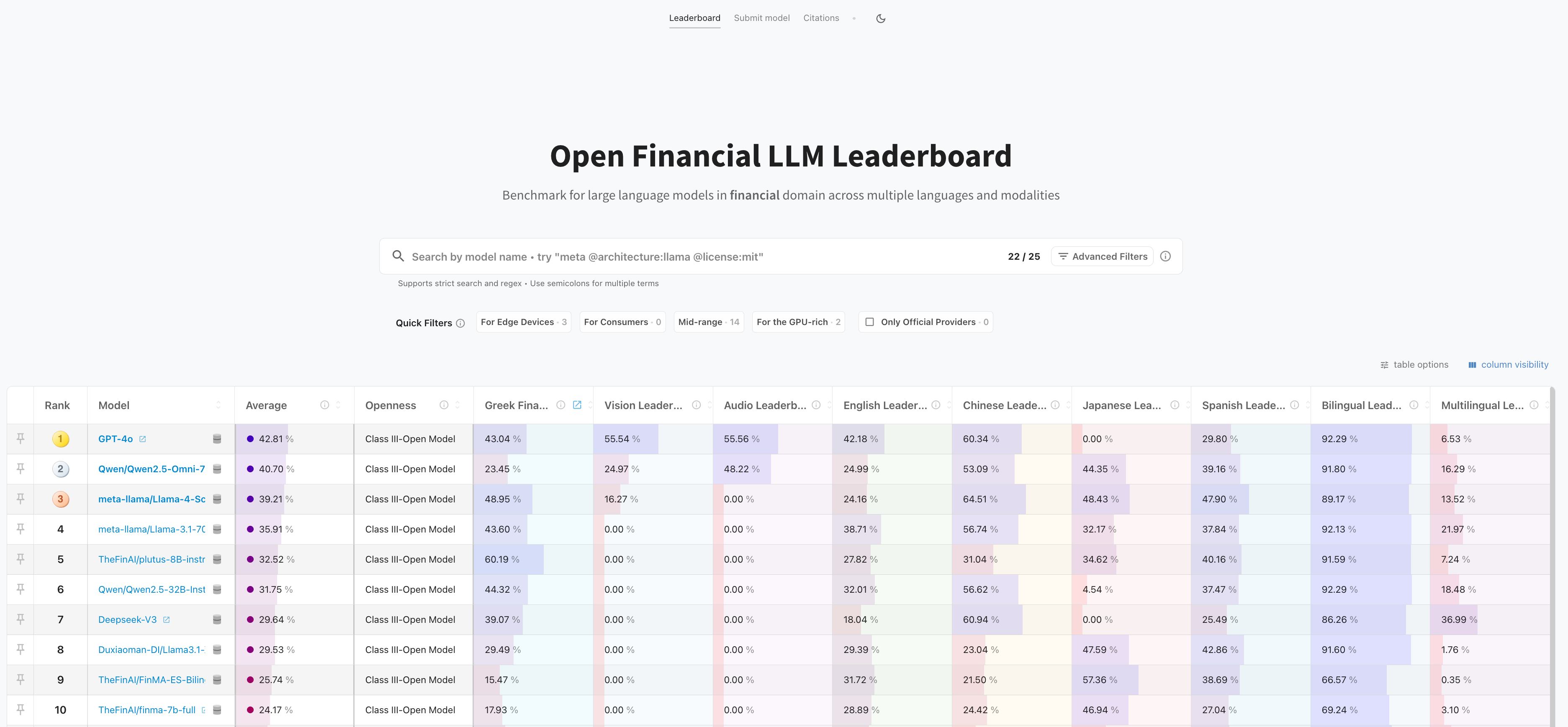}
  \caption{The \textsc{\multifinben} leaderboard.}
  \label{fig:fig_leaderboard}
\end{figure}

To promote transparency, openness, and community engagement, we have developed an interactive leaderboard for \textsc{\multifinben} (Figure~\ref{fig:fig_leaderboard}). The leaderboard displays model performance across all benchmark tasks and includes Model Openness Framework (MOF) \cite{mof2024white} tags, which provide structured metadata about each model’s openness, accessibility, and reproducibility. Users can filter, compare, and explore submissions based on performance metrics and openness levels. The leaderboard is continuously updated and publicly accessible to encourage standardized and responsible evaluation practices in financial AI.
% \footnote{\url{https://huggingface.co/spaces/TheFinAI/Open-FinLLM-Leaderboard}}

% \section{Author Contribution}

% The author contributions are summarized below:

% \begin{itemize}
%     \item \textbf{Science Leadership}:  Kaleb E Smith, Meikang Qiu, Arman Cohan, Xiao-Yang Liu, Jimin Huang, Guojun Xiong, Alejandro Lopez-Lira, Xi Chen, Junichi Tsujii, Jian-Yun Nie, Sophia Ananiadou, Qianqian Xie
%     \item \textbf{Contributors}: Xueqing Peng, Lingfei Qian, Yan Wang, Ruoyu Xiang, Yueru He, Yang Ren, Mingyang Jiang, Vincent Jim Zhang, Yuqing Guo, Jeff Zhao, Huan He, Yi Han, Yun Feng, Yuechen Jiang, Yupeng Cao, Haohang Li, Yangyang Yu, Xiaoyu Wang, Penglei Gao, Shengyuan Lin, Keyi Wang, Shanshan Yang, Yilun Zhao, Zhiwei Liu, Peng Lu, Jerry Huang, Suyuchen Wang, Triantafillos Papadopoulos, Polydoros Giannouris, Efstathia Soufleri, Nuo Chen, Zhiyang Deng, Heming Fu, Yijia Zhao, Mingquan Lin
% \end{itemize}

\end{document}